%% file: main.tex
\documentclass{article}
\input{math_commands}

\usepackage{microtype}
\usepackage{graphicx}
\usepackage{subfig}
\usepackage{booktabs} 

\usepackage{hyperref}

\usepackage[preprint]{icml2026}

\usepackage{amsmath}
\usepackage{amssymb}
\usepackage{mathtools}
\usepackage{amsthm}
\usepackage{siunitx}
\usepackage{enumitem}

\usepackage[capitalize,noabbrev]{cleveref}

\theoremstyle{plain}

\theoremstyle{definition}

\theoremstyle{remark}

\usepackage[textsize=tiny]{todonotes}

\icmltitlerunning{An Overview of Prototype Formulations for Interpretable Deep Learning}

\begin{document}

\twocolumn[
\icmltitle{An Overview of Prototype Formulations for Interpretable Deep Learning}



\icmlsetsymbol{equal}{*}

\begin{icmlauthorlist}
\icmlauthor{Maximilian Xiling Li}{kit}
\icmlauthor{Korbinian Rudolf}{kit}
\icmlauthor{Paul Mattes}{kit}
\icmlauthor{Nils Blank}{kit}
\icmlauthor{Rudolf Lioutikov}{kit}
\end{icmlauthorlist}

\icmlaffiliation{kit}{Intuitive Robots Lab, Karlsruhe Institute of Technology, Karlsruhe, Germany}

\icmlcorrespondingauthor{Maximilian X. Li}{maximilian.li@kit.edu}
\icmlcorrespondingauthor{Rudolf Lioutikov}{\mbox{lioutikov@kit.edu}}

\icmlkeywords{XAI, Prototype Learning, Image Classification, Deep Learning}

\vskip 0.3in
]



\printAffiliationsAndNotice{}  

\newcommand{\ournet}{HyperPGNet}
\newcommand{\cvec}[1]{\boldsymbol{{#1}}}
\newcommand{\cmat}[1]{\boldsymbol{\mathrm{#1}}}

\newcommand{\anchor}{\cvec{\alpha}}

\input{content/00_abstract}
\input{content/01_introduction}

\input{content/02_relwork}

\input{content/03_hyperpg}
\input{content/04_training}
\input{content/05_experiments}

\input{content/06-xai}
\input{content/06_conclusion}

\section*{Acknowledgements}
The research presented in this paper was funded by the Deutsche Forschungsgemeinschaft (DFG, German
Research Foundation) – 448648559.

\section*{Impact Statement}

This paper presents work whose goal is to advance the field of 
Machine Learning. There are many potential societal consequences 
of our work, none which we feel must be specifically highlighted here.

\bibliography{main}
\bibliographystyle{icml2026}

\newpage
\appendix
\onecolumn
\input{appendix/app-xai}
\input{appendix/app-implementation}
\input{content/05a_ablations}
\input{appendix/app-hyperpg-distr}
\input{appendix/app-scaleddot}
\input{appendix/app-hyperpg-params}

\end{document}

%% file: math_commands.tex
\usepackage{amsmath,amsfonts,bm}

\def\eqref#1{equation~\ref{#1}}

\def\1{\bm{1}}

\def\vp{{\bm{p}}}

\def\vv{{\bm{v}}}

\def\vx{{\bm{x}}}

\def\vz{{\bm{z}}}

\def\mA{{\bm{A}}}

\def\mI{{\bm{I}}}

\def\mK{{\bm{K}}}

\def\mP{{\bm{P}}}
\def\mQ{{\bm{Q}}}

\def\mV{{\bm{V}}}

\def\mX{{\bm{X}}}
\def\mY{{\bm{Y}}}
\def\mZ{{\bm{Z}}}

\DeclareMathAlphabet{\mathsfit}{\encodingdefault}{\sfdefault}{m}{sl}
\SetMathAlphabet{\mathsfit}{bold}{\encodingdefault}{\sfdefault}{bx}{n}

\def\gN{{\mathcal{N}}}



%% file: content/00_abstract.tex
\begin{abstract}
Prototypical part networks offer interpretable alternatives to black-box deep learning models by learning visual prototypes for classification.
This work provides a comprehensive analysis of prototype formulations, comparing point-based and probabilistic approaches in both Euclidean and hyperspherical latent spaces. 
We introduce HyperPG, a probabilistic prototype representation using Gaussian distributions on hyperspheres. Experiments on CUB-200-2011, Stanford Cars, and Oxford Flowers datasets show that hyperspherical prototypes outperform standard Euclidean formulations. 
Critically, hyperspherical prototypes maintain competitive performance under simplified training schemes, while Euclidean prototypes require extensive hyperparameter tuning.
\end{abstract}

%% file: content/01_introduction.tex
\section{Introduction}
Deep Learning has achieved high accuracy in many computer vision tasks. However, the decision-making processes of these models lack transparency and interpretability, making deployment in safety-critical areas challenging. Explainable Artificial Intelligence (XAI) seeks to develop interpretability methods to open the black-box reasoning processes of models and increase trust in their decisions.

XAI methods can be broadly divided into two categories: First, post-hoc methods like LIME \cite{Ribeiro2016WhyShouldI}, SHAP \cite{Lundberg2017UnifiedApproachInterpreting} or GradCAM \cite{GradCam}, which explain model predictions without retraining. 
While broadly applicable, post-hoc methods may not align with the models' actual decision-making processes \cite{Rudin2019Stopexplainingblack}. 
Second, inherently interpretable methods provide built-in, case-based reasoning. For instance, small decision trees are inherently interpretable through their if-else structure \cite{Molnar2020Interpretablemachinelearning}, but are limited in representational power.


\begin{figure}
    \centering
    \includegraphics[width=0.9\linewidth]{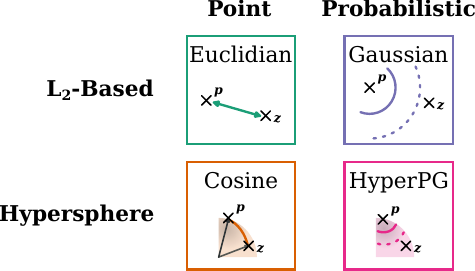}
    \caption{Different Prototype Formulations. HyperPG is a novel formulation for probabilistic prototypes on a Hypersphere.}
    \label{fig:tablefig}
\end{figure}

Deep prototype learning architectures such as ProtoPNet \cite{Chen2019ThisLooksThat} and its derivatives \cite{Rymarczyk2020ProtopsharePrototypesharing, Donnelly2021DeformableProtoPNetInterpretable, Sacha2023ProtosegInterpretablesemantic} integrate inherent interpretability into deep learning through a prototype layer.
Each prototype neuron stores a latent feature vector, with predictions based on distances between sample features and prototypes. This makes prototype learning a clustering task in latent space \cite{Zhou2022RethinkingSemanticSegmentation}. 
Different formulations induce different latent structures: cosine similarity induces hyperspherical structure with classification advantages \cite{Mettes2019HypersphericalPrototypeNetworks}, while probabilistic formulations like Gaussian prototypes enable Bayesian downstream tasks. 

\input{tab/tab_overview}

The increased popularity of prototypical part models increases architectural choices, yet there is no comprehensive overview of prototype formulations and their effect on predictive performance. This paper explores different prototype formulations with the following contributions:
\begin{itemize}[leftmargin=*,topsep=2pt,itemsep=1pt]
    \item \textbf{Prototype Formulation Overview:} A comprehensive overview of different learnable prototype definitions, including both point-based and probabilistic prototypes as shown in \autoref{fig:tablefig}.
    \item \textbf{HyperPG:} A probabilistic prototype representation modeling Gaussian distributions over cosine similarities with learned anchor $\alpha$, mean $\mu$, and variance $\sigma^2$, projecting Gaussians onto hypersphere surfaces.
    \item \textbf{Benchmarking and Robustness Analysis:} Image classification experiments on CUB-200-2011 \cite{Wah2011caltechucsdbirds}, Stanford Cars \cite{Krause}, and Oxford Flowers \cite{Nilsback2008AutomatedFlower} datasets,     evaluating both full optimization and simplified training procedures to reveal     differences in hyperparameter sensitivity. Our code is publicly available on Github\footnote{\url{https://github.com/LiXiling/prob-proto}}.
\end{itemize}

%% file: tab/tab_overview.tex
\begin{table*}[t]
\centering
\caption{Overview of 15 related models and their prototype learning configurations.}
\label{tab:works_overview}
\vskip 0.15in
\resizebox{\textwidth}{!}{%
\begin{tabular}{@{}lllllr@{}}
\toprule
\textbf{Model}     & \textbf{Similarity} & \textbf{Shape}         & \textbf{Assignment}  & \textbf{Clf. Head}  &         \\ \midrule
Prototype Decoder       & Euclidean  & Entire Image  & Class Exclusive       & FCL                 & \cite{Li2018Deeplearningcase}\\
ProtoPNet               & Euclidean  & Patch         & Class Exclusive       & FCL                 & \cite{Chen2019ThisLooksThat}\\
Def. ProtoPNet & \textbf{Cosine}    & Spatial Arrangement      & Class Exclusive & FCL                & \cite{Donnelly2021DeformableProtoPNetInterpretable} \\
ProtoPShare             & Euclidean  & Patch         & Merged after Training & FCL                 & \cite{Rymarczyk2020ProtopsharePrototypesharing} \\
ProtoPool               & Euclidean  & Patch Pooling & Shared                & FCL                 & \cite{Rymarczyk2022Interpretableimageclassification} \\
TesNet                  & \textbf{Grassman}   & Patch         & Class Exclusive       & FCL                 & \cite{Wang2021TesNet}\\
ProtoTree               & Euclidean  & Patch         & Shared                & Decision Tree       & \cite{Nauta2021NeuralPrototypeTrees}\\
ProtoKNN                & \textbf{Cosine}     & Patch         & Class Exclusive       & KNN Clf             & \cite{Ukai2023ThisLooksIt}\\
PIPNet                  & \textbf{Cosine}     & Patch         & Shared                & FCL                 & \cite{Nauta2023PIPNetPatch}\\
ST-ProtoPNet & \textbf{Cosine} & Patch & Class Exclusive & Branch Aggregation & \cite{wang2023learning} \\
LucidPPN             & Euclidean & Separate Color \& Texture & Class Exclusive & Branch Aggregation & \cite{Pach2024LucidPPNUnambiguousPrototypical}\\
ProtoSeg & \textbf{Cosine}     & Patch         & Class Exclusive       & FCL                 & \cite{Zhou2022RethinkingSemanticSegmentation}\\
ProtoGMM                & \textbf{Gaussian}   & Patch         & Class Exclusive       & FCL                 & \cite{Moradinasab2024ProtoGMMMultiprototype}\\
MGProto                 & \textbf{Gaussian}   & Patch         & Class Exclusive       & Bayesian Likelihood & \cite{Wang2024MixtureGaussianDistributed}\\ 
ProtoPFormer & Euclidean & Transformer Token & Class Exclusive & Branch Aggregration & \cite{Xue2024ProtoPFormer}\\
\bottomrule
\end{tabular}
}
\end{table*}

%% file: content/02_relwork.tex
\section{Related Work}

\textbf{Prototype Learning.} In image classification, prototype learning approaches using autoencoders provide high interpretability by reconstructing learned prototypes from latent space back to the image space \citep{Li2018Deeplearningcase}. However, these approaches are limited in their performance because each prototype must represent the entire image. ProtoPNet \citep{Chen2019ThisLooksThat} introduced the idea of prototypical parts. In this setting, each prototype is a latent patch of the input image, commonly a $1 \times 1$ latent patch. The prototypes are each associated with a single class and learned via backpropagation without additional information. The similarity of the prototypes to the image patch is based on the Euclidean distance.

Multiple successors build on the idea of ProtoPNet. ProtoPShare \citep{Rymarczyk2020ProtopsharePrototypesharing} merges the class-exclusive prototypes to class-shared ones. ProtoPool \citep{Rymarczyk2022Interpretableimageclassification} directly learns class-shared patch prototypes and pools them by learning a slot assignment. Deformable ProtoPNet \citep{Donnelly2021DeformableProtoPNetInterpretable} learns a mixture of prototypical parts with dynamic spatial arrangement. For computing the similarity between the prototypes and image patches, Deformable ProtoPNet uses the cosine similarity. TesNet \citep{Wang2021TesNet} proposes to compute the prototype similarity on the Grassman Manifold.

\begin{figure*}[ht!]
    \centering
    \includegraphics[width=0.85\linewidth]{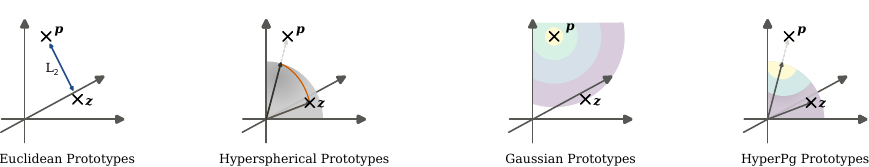}
    \caption{Illustration of similarity computation between prototype $\vp$ and latent vector $\vz$ for different formulations. 
    Euclidean: $L_2$ distance in latent space. 
Hyperspherical: cosine similarity of normalized vectors (angle on hypersphere). 
Gaussian: PDF of Gaussian distribution in Euclidean space. 
HyperPG: PDF of Gaussian distribution over cosine similarities (Gaussian on hypersphere surface).}
    \label{fig:prototypes-overview}
\end{figure*}

All these approaches share the use of a linear classification layer. However, ProtoTree \cite{Nauta2021NeuralPrototypeTrees} builds a Decision Tree on the learned Euclidean prototypes and ProtoKNN \cite{Ukai2023ThisLooksIt} proposes to use a $k$-nearest-neighbor classifier with cosine prototype similarities. ST-ProtoPNet \cite{wang2023learning} learn the prototypes close to the decision boundary, similar to support vectors.

Recent work proposes to learn more interesting or robust features for the prototypes from the image encoders. PIPNet \citep{Nauta2023PIPNetPatch} use augmentations during training to align the learned prototypes with more meaningfull image content. LucidPPN \citep{Pach2024LucidPPNUnambiguousPrototypical} propose a hybrid architecture with texture only grayscale prototypes and low resolution color only prototypes.

``Rethinking Semantic Segmentation'' \cite{Zhou2022RethinkingSemanticSegmentation} view prototype learning as a clustering task in the network's latent space. The clustering is done outside of the network's backpropagation path with an external clustering algorithm, so-called \textit{non-learnable prototypes}. Later work such as ProtoGMM \cite{Moradinasab2024ProtoGMMMultiprototype} and MGProto \cite{Wang2024MixtureGaussianDistributed} build on this idea and model the prototypes as Gaussian distributions.

ProtoPFormer \cite{Xue2024ProtoPFormer} adapts Euclidean ProtoPNets to Transformers through a ViT \cite{Dosovitskiy2020ImageIsWorth} backbone. To avoid the distraction problem due to the global attention mechanism of ViTs, ProtoPFormer uses a hybrid approach with global and local prototypes. The local prototypes are encouraged to focus on different parts of the image by modeling the \textit{spatial} prototype activations as a 2D Gaussian.

\autoref{tab:works_overview} presents a systematic overview, where the cosine similarity performs well in classification tasks \cite{Mettes2019HypersphericalPrototypeNetworks}, but only few prototype learning works have used them so far \cite{Donnelly2021DeformableProtoPNetInterpretable, Zhou2022RethinkingSemanticSegmentation, Ukai2023ThisLooksIt}. 
There is a lack for probabilistic prototypes in a hyperspherical space. We introduce HyperPG, that combines the performance gains from the cosine similarity with the probabilistic nature of Gaussian distributions.

%% file: content/03_hyperpg.tex
\section{Overview: Prototype Formulations}

Prototype Learning is an inherently interpretable machine learning method. The reasoning process is based on the similarity scores of the inputs to the prototypes, retained representations of the training data. For example, a K-Nearest Neighbor (KNN) model is a prototype learning approach with the identity function for representation and an unlimited number of prototypes. In contrast, a Gaussian Mixture Model (GMM) uses a mean representation but restricts the number of prototypes to the number of mixture components.

Prototype learning for deep neural networks involves finding structures in latent space representations. This section provides an overview of existing methods, which are also illustrated in \autoref{fig:prototypes-overview}. Prior work uses point-based prototypes, computing similarity scores relative to a single point in latent space. On the other hand, probabilistic formulations such as Gaussian prototypes or HyperPG prototypes allow the model to adapt to the variance in the training data.

\subsection{Point Based Prototypes}
The general formulation of prototypes, as defined in previous work \citep[e.g.,][]{Chen2019ThisLooksThat}, is discussed first.
Let $\mathcal{D} = [\mX, \mY] = \{(\vx_i, y_i)\}_{i=1}^N$ denote the training set, e.g., a set of labeled images, with classes $C$. 
Each class $c \in C$ is represented by $Q$ many prototypes 
$\mP_c = \{\vp_{c,j}\}_{j=1}^Q$.

Some feature encoder $\mathrm{Enc}$ projects the inputs into a $D$-dimensional latent space $\mathcal{Z}$, with \\$\vz_i = \mathrm{Enc}(\vx_i)$ being a feature map of shape $\zeta_w \times \zeta_h \times D$ with spatial size $\zeta = \zeta_w\zeta_h$. 
Commonly, the prototypes $\vp$ are also part of $\mathcal{Z}$ with shape $\rho_w \times \rho_h \times D$, i.e., spatial size $\rho = \rho_w \rho_h$. 

Early approaches based on Autoencoder architectures \cite{Li2018Deeplearningcase} use $\rho=\zeta$, meaning the prototype represents the entire image and can therefore be decoded out from latent space back to the input space. Part-based approaches like ProtoPNet and segmentation models like ProtoSeg use $\rho=1$ \citep{Chen2019ThisLooksThat, Zhou2022RethinkingSemanticSegmentation}, meaning each prototype represents some part of the image.
Notable exceptions include Deformable ProtoPNet \citep{Donnelly2021DeformableProtoPNetInterpretable}, where each prototype has is a $\rho = 3 \times 3$ arrangement of smaller patches, and MCPNet \citep{Wang2024MCPNetInterpretableClassifier}, which learns concept prototypes across the latent features.

The prediction is done by comparing each prototype $\vp$ to the latent feature map $\vz$. For simplicity, lets assume the spatial dimensions $\rho=\zeta=1$. The following equations can be adapted for higher spatial dimensions by summing over the height and width $\sum_{\rho_w} \sum_{\rho_h}$ for each patch of the latent map.

\subsubsection{Euclidean Prototypes}

ProtoPNet's Euclidean prototypes leverage the $L_2$ similarity. The $L_2$ similarity measure is defined as
\begin{equation}
\label{eq:l2sim}
    s_{L_2}(\vz | \vp) = 
        \log \left(
            \frac{\| \vz - \vp \|^2_2 + 1
            }{\| \vz - \vp \|^2_2 + \epsilon}\right)
\end{equation}
and is based on the inverted $L_2$ distance between a latent vector $\vz$ and a prototype vector $\vp$. This similarity measure starts to perform worse with higher numbers of dimensions, as in a large enough space all points are distant to each other and the $L_2$ distance looses meaning.

\subsubsection{Cosine Prototypes}
\newcommand{\scos}{s_{\cos}(\vz | \cvec{\alpha})}
Prototype formulations based on the cosine similarity create a hyperspherical space. They have been shown to perform well in classification tasks \cite{Mettes2019HypersphericalPrototypeNetworks} and recent works have started to apply them \citep[e.g.,][]{Donnelly2021DeformableProtoPNetInterpretable, Zhou2022RethinkingSemanticSegmentation, Nauta2021NeuralPrototypeTrees, Ukai2023ThisLooksIt}.
The cosine similarity is defined as
\begin{equation}
\label{eq:cossim}
    s_{\cos} (\vz | \vp) = \frac{\vz^\top \vp}{\| \vz \|_2\| \vp\|_2},
\end{equation}
which is based on the angle between two normalized vectors of unit length. By normalizing $D$ dimensional vectors to unit length, they are projected onto the surface of a $D$ dimensional hypershere. The cosine similarity is defined on the interval $[-1, 1]$ and measures: $1$ for two vectors pointing in the same direction, $0$ for orthogonal vectors, and $-1$ for vectors pointing in opposite directions.  Without the normalization to unit length, we can define a scaled dot-product prototype inspired by the Attention mechanism \cite{Vaswani2017Attention}. However, this similarity measure is not bounded to any interval, as the vector length is proportionally preserved. For more details see \autoref{app:scaleddot}. Like the $L_2$ similarity, the cosine similarity is a point-based measure comparing two vectors directly.

Both the $L_2$ and cosine similarity have been used for classification tasks. The similarity scores are processed by a fully connected layer \citep[e.g.,][]{Chen2019ThisLooksThat, Donnelly2021DeformableProtoPNetInterpretable}, or a winner-takes-all approach assigns the class of the most similar prototype \citep[e.g.,][]{Sacha2023ProtosegInterpretablesemantic}. The prototypes can be learned by optimizing a task-specific loss, such as cross-entropy, via backpropagation. Alternatively, \citet{Zhou2022RethinkingSemanticSegmentation} propose ``non-learnable'' prototypes, whose parameters are obtained via a clustering operation in the latent space instead of backpropagation.

\subsection{Probabilistic Prototypes}
Viewing prototype learning as clustering problem, point-based prototypes compute the similarity of a latent vector to the center coordinate of each cluster, as represented by the prototype vector. Probabilistic prototypes aim to model the cluster as a probability distribution. Future work could use probabilistic prototypes for extended downstream tasks such as outlier detection or interventions by sampling from generative distributions.

\subsubsection{Gaussian Prototypes}
Gaussian prototypes model the clusters in latent space as a Gaussian distribution with mean and covariance. By adapting the covariance matrix to the training data a Gaussian prototype with a wide covariance can still have a relatively high response even for larger $L_2$ distances from the mean vector. On the other hand, with a very small covariance matrix a Gaussian prototype could show no response unless the mean value is met nearly exactly.

Let the formal definition of a Gaussian prototype be $\vp^{G} = (\cvec{\mu}, \cmat{\Sigma})$. The parameters of $\vp^G_{c,j}$ now track both the mean and covariance of latent vector distribution $\mZ_{c,j}$. 
Each Gaussian prototype $\vp^G$ thus defines a multivariate Gaussian Distribution $\gN(\cvec{\mu}, \cmat{\Sigma})$. Gaussian prototypes can be trained using EM for clustering in the latent space \citep{Zhou2022RethinkingSemanticSegmentation, Moradinasab2024ProtoGMMMultiprototype, Wang2024MixtureGaussianDistributed} or naively by directly optimizing the parameters via Backpropagation. 
The similarity measure of Gaussian prototypes is defined as the probability density function (PDF) for $D$-dimensional multivariate Gaussians, namely
\begin{align}
    s_\mathrm{Gauss}(\vz | \vp^G) &= \mathcal{N}(\vz ; \cvec{\mu}, \cmat{\sigma})
    \\
    = \frac{1}{(2\pi)^{\frac{D}{2}} |\cmat{\Sigma}|^{\frac{1}{2}}} &
    \exp \left(- \frac{1}{2} 
       (\vz - \cvec{\mu})^\top \cmat{\Sigma}^{-1} (\vz - \cvec{\mu})        
    \right).
\end{align}
The term inside the exponent is based on the $L_2$ distance between the latent vector $\vz$ and the mean $\cvec{\mu}$.
The PDF formulation has several advantages: A) The similarity can be interpreted as the likelihood of being sampled from the Gaussian prototypes, which is more meaningful than a distance metric in a high-dimensional latent space. B) Prototypes can adapt their shape using a full covariance matrix, allowing different variances along various feature dimensions, offering more flexibility in shaping the latent space.
However, learning a full covariance matrix increases computational requirements.
\begin{figure}[t]
    \centering
    \subfloat[$\mu=1$]{
        \includegraphics[width=0.3\linewidth]{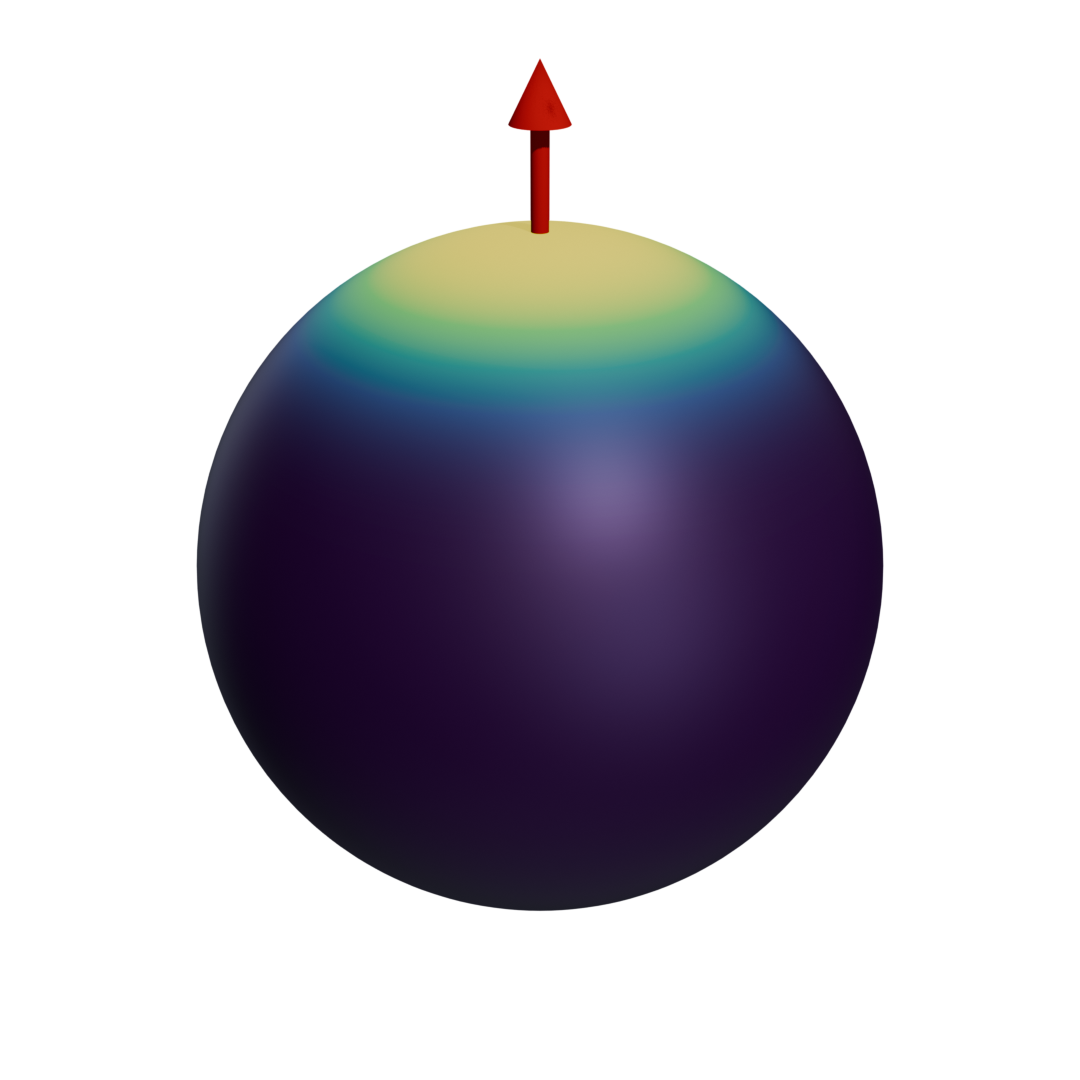}
    }
    \subfloat[$\mu=0.5$]{
        \includegraphics[width=0.3\linewidth]{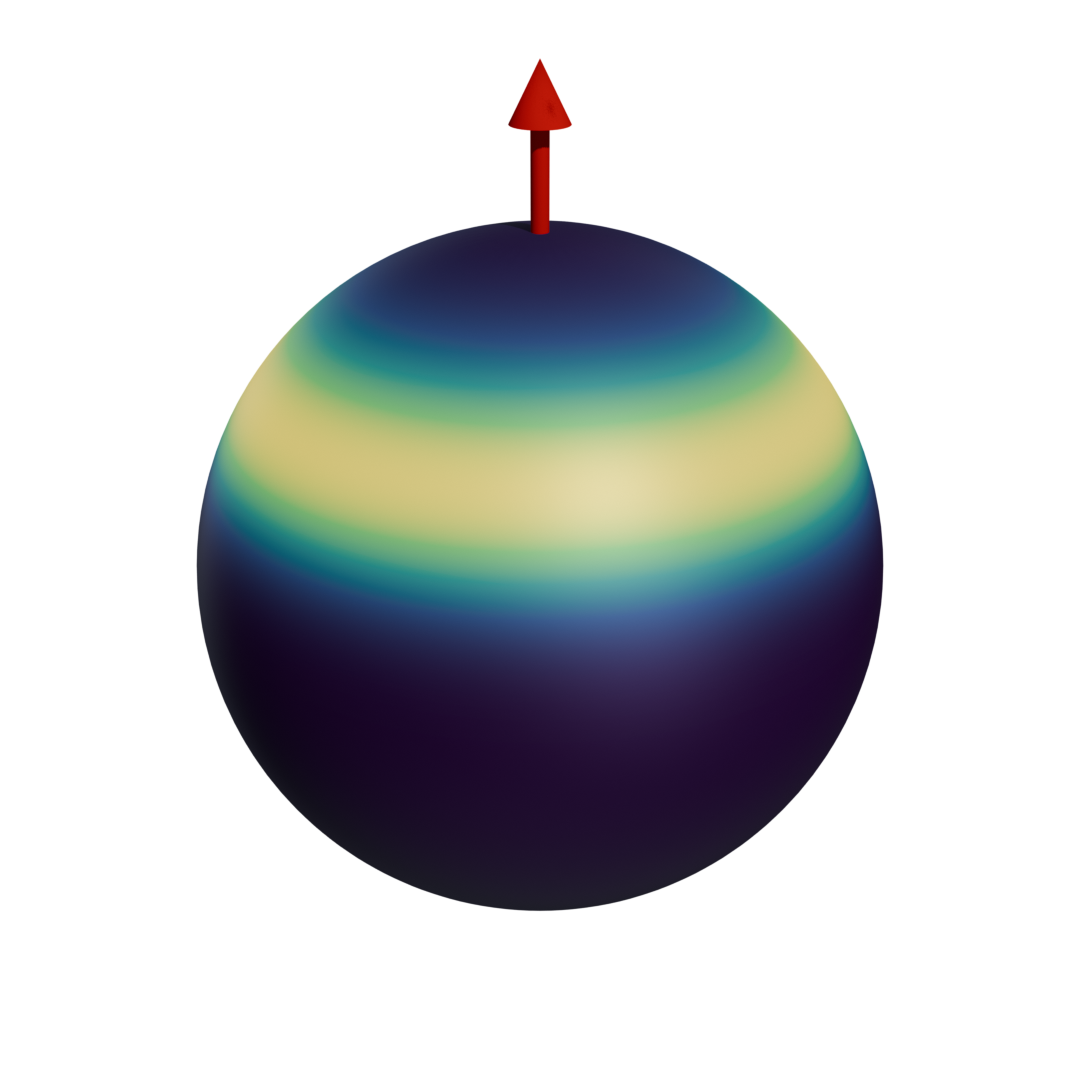}
    }
    \subfloat[$\mu=0$]{
        \includegraphics[width=0.3\linewidth]{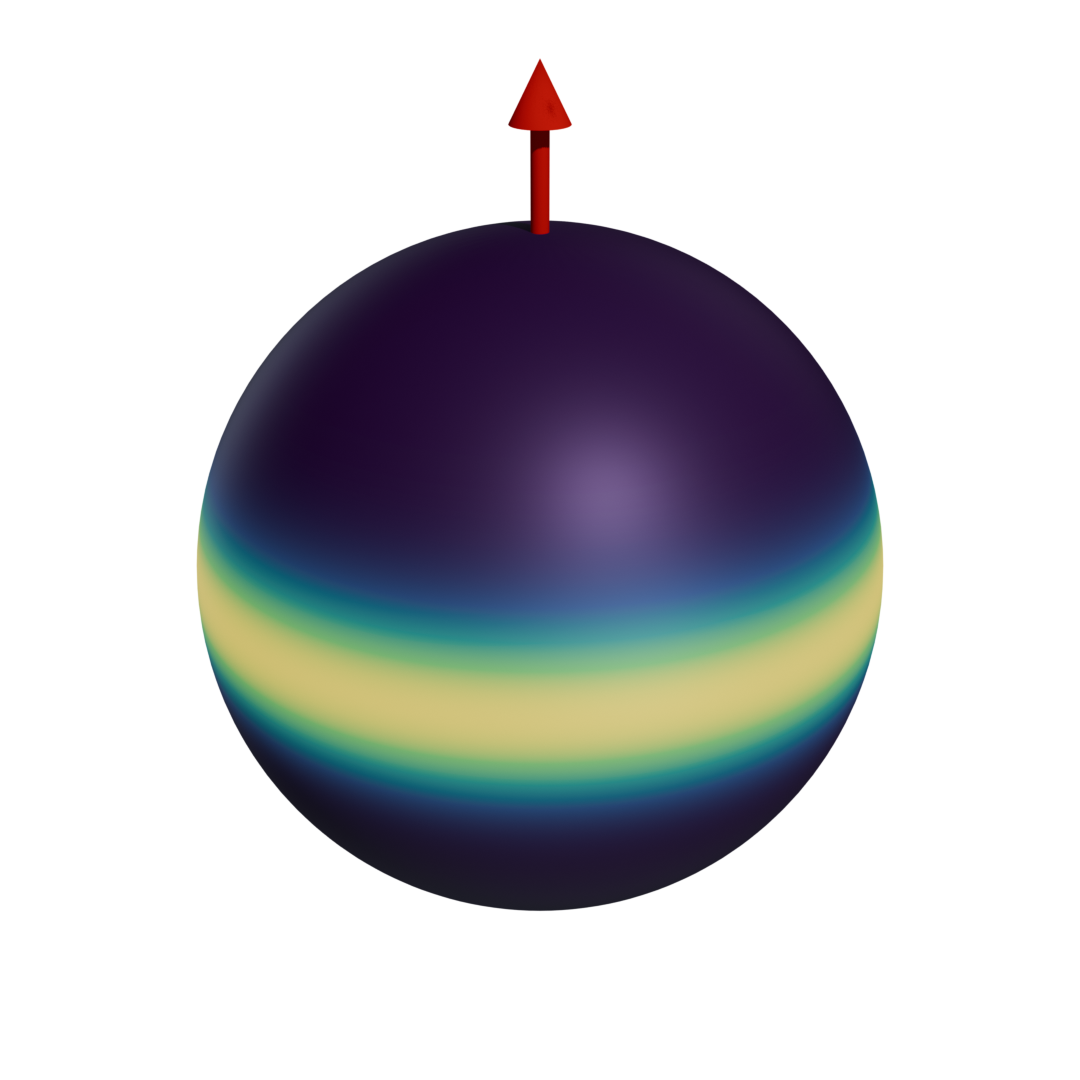}
    }
    \caption{HyperPG prototypes learn a distribution of cosine similarities. They use a learnable anchor vector $\anchor$, scalar mean $\mu$ and variance $\sigma^2$. They project a Gaussian distribution of cosine similarities on the surface of a hypersphere, resulting in ring shaped activation patterns around the anchor vector.}
    \label{fig:hyperpg-spheres}
\end{figure}

\begin{figure*}
    \centering
    \includegraphics[width=0.8\linewidth]{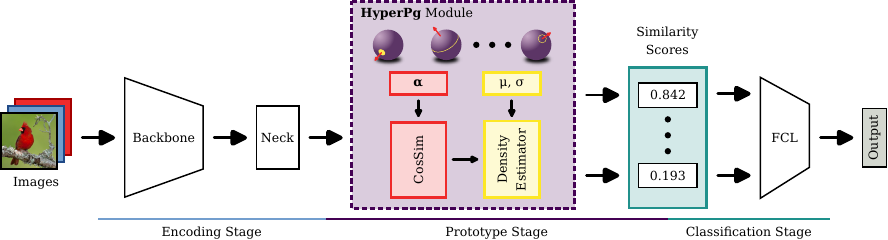}
    \caption{Prototype Learning Architecture. The HyperPG module can be easily exchanged to other prototype formulations such as Euclidian or Cosine prototypes. HyperPG uses a Gaussian distribution as Density Estimator, but other PDFs are possible.}
    \label{fig:architecture}
\end{figure*}

\subsubsection{Gaussian Prototypes on the Hypersphere - HyperPG}
\label{ssec:hyperpg-def}
Just as Gaussian prototypes provide a probabilistic formulation for Euclidean prototypes, we aim to develop a probabilistic formulation for hyperspherical prototypes based on the cosine similarity. We name this new formulation \textit{Prototypical Gaussians on the Hypersphere} (HyperPG). HyperPG prototypes are defined as $\vp^H = (\cvec{\alpha}, \mu, \sigma)$ with a directional anchor vector $\cvec{\alpha}$, \textit{scalar} mean similarity $\mu$ and \textit{scalar} standard deviation (std) $\sigma$. HyperPG prototypes learn a $1D$ Gaussian distribution over the cosine similarities to the anchor vector $\cvec{\alpha}$. Because the cosine similarity is bounded to $[-1,1]$, HyperPG's similarity measure is defined as the PDF of the truncated Gaussian distribution within these bounds. Let $\mathcal{G}(x, \mu, \sigma)$ be the cumulative Gaussian distribution function. Then, HyperPG's similarity measure based on the truncated Gaussian distribution is defined as
\begin{align}
    s_\mathrm{HyperPG}(\vz | \vp^H) 
    &= \mathcal{T}_G(\scos; \mu, \sigma, -1, 1)
    \\
    &= \frac{
        \gN\left(
            \scos; \mu, \sigma
        \right)
        }{
            \mathcal{G}(1, \mu, \sigma) - \mathcal{G}(-1, \mu, \sigma)            
        }.
\end{align}

\autoref{fig:hyperpg-spheres} illustrates the activations of HyperPG's similarity function on the surface of a $3D$ hypersphere with anchor $\cvec{\alpha} = (0, 0, 1)$, fixed std $\sigma=0.1$ and various mean values $\mu \in [0,1]$. The anchor $\cvec{\alpha}$ defines a prototypical direction vector in latent space $\mathcal{Z}$, similar to other hyperspherical prototypes, and is visualized as a red arrow. The learned Gaussian distribution of cosine similarities is projected onto the hypersphere's surface with std $\sigma$ governing the spread of the distribution, and mean $\mu$ the expected distance to the anchor $\cvec{\alpha}$. For $\mu = 1$, the distribution centers around the anchor, as the cosine similarity is $1$ if two vectors point in the same direction. This corresponds to the \textit{von Mises-Fisher} (vMF) distribution (\autoref{app:hyperpg-ext}).
For $\mu=-1$, the distribution is the on opposite side of the hypersphere, as the cosine similarity is $-1$ for vectors pointing in opposite directions. 

For values of $1 > \mu > -1$, the distribution forms a hollow ring around the anchor vector $\cvec{\alpha}$. 
This occurs because the cosine similarity for these $\mu$ values expects the activating vectors to point in a different direction than the anchor, without specifying the direction. Imagining the interpolation between $\mu =1$ and $\mu = -1$, the probability mass moves from one pole of the hypersphere to the other, stretching like a rubber band over the surface. For $\mu=0$, the expected cosine similarity indicates that vectors with the highest activation are orthogonal to the anchor $\cvec{\alpha}$. Since no specific direction is indicated, the entire hyperplanar segment orthogonal to the anchor has the highest activation. 

These ring-like activation patterns would require an infinite mixture of prototype vectors pointing in all directions in this hyperplane. HyperPG achieves the same effect by learning only one prototype vector (the anchor) and just two additional scalar parameters. This significantly increases HyperPG's representational power compared to Gaussian prototypes without increasing the computational complexity. This potential capability is also a major difference to the vMF distribution.

One drawback of the HyperPG formulation, is that it cannot be ``pushed'' to the nearest training sample as with the other formulations. However, the prototype's interpretability can be maintained by following the approach of prototypical concept balls \cite{ma2023this}. Instead of only using the pushed training sample to interpret the prototype, all training sample with similar activation are equally good explanations. By presenting multiple sample patches the user is able to infer a learned concept.

HyperPG can be easily adapted to other probability distributions with additional desirable properties. Possible candidate distributions are elaborated on in \autoref{app:hyperpg-ext}. Similarly, it is possible to exchange the cosine similarity to other similarity measures or functions, and learn an untruncated PDF over their output, making the HyperPG idea transferable to other manifolds and applications outside of prototype learning.

%% file: content/04_training.tex
\section{Training Prototypical Part Networks}
\label{sec:training}
Prototypical Part networks like ProtoPNet have a reputation for being difficult to tune and hyperparameter-sensitive. To better explore the design choices, we categorize them into three groups: (1) Network architecture governs information flow from input to the output, including pretrained backbone network and neck for feature extraction. (2) Prototype hyperparameters shape the prototypical latent space through the choice of prototype formulation, but also number of prototypes, prototype dimensions and additional shaping losses. (3) Network optimization settings such as warmup-epochs, learning rate schedules or number of optimizers. ProtoPNet and its successors often rely on complex optimization schemes with warm-up phases and stage-specific optimizers and schedules. Detailed hyperparameter values are provided in \autoref{app:hyperparam}. 

\subsection{Network Architecture}
For our experiments, we use the ProtoPNet's architecture \cite{Chen2019ThisLooksThat} as the foundational deep prototype learning model for image classification.
\autoref{fig:architecture} illustrates this network architecture with the HyperPG prototype module. The encoding stage employs a pretrained backbone (e.g., ResNet \cite{He2016Deepresiduallearning}) . A subsequent projection neck consisting of two $1 \times 1$ convolution layers with ReLU between them and Sigmoid activation at the end, reduce dimensionality to create the latent prototype space.

In the prototype stage, similarity scores are computed. Original ProtoPNet uses Euclidean prototypes measuring $L_2$ distance between encoded features and prototype parameters. The HyperPG module, shown in \autoref{fig:architecture}, first calculates cosine similarity between learnable anchors $\cvec{\alpha}$ and neck-produced latent vectors. Then, a Density Estimation layer with parameters $\mu$ and $\sigma$ computes Gaussian PDF over the previous layer's activations. Both components could be independently modified in future work, implementing alternative similarity measures (e.g., hyperbolic) or multi-modal probability distributions.

The classification stage produces output logits from similarity scores through a single linear layer, preserving interpretability.

\subsubsection{Comparison to Recent Works}
Our experiments use the basic ProtoPNet architecture, focusing on the effect of different prototype formulations. This section briefly outlines modifications required to implement the models in \autoref{tab:works_overview}.
LucidPPN \cite{Pach2024LucidPPNUnambiguousPrototypical} replaces the encoding stage with a hybrid encoder for color-only and texture-only feature extraction. Deformable ProtoPNet \cite{Donnelly2021DeformableProtoPNetInterpretable} modifies the prototype module's input to allow dynamic spatial arrangements (``deformations'').  ProtoPool \cite{Rymarczyk2022Interpretableimageclassification} alters the prototype module's output by learning to dynamically pool prototype activations. ProtoTree \cite{Nauta2021NeuralPrototypeTrees} and ProtoKNN \cite{Ukai2023ThisLooksIt} use a different inherently interpretable models for the classification stage. ProtoPFormer \cite{Xue2024ProtoPFormer} utilizes a Vision Transformer backbone with separate network branches for global information (class token) and local information (image tokens).

\subsection{Prototype Losses}
The original ProtoPNet implementation uses three loss functions: a task specific loss like crossentropy for classification, a cluster loss to increase compactness within a class's cluster, and a separation loss to increase distances between different prototype clusters.

ProtoPNet defines a cluster loss function to shape the latent space such that all latent vectors $\vz_c \in \mZ_c$ with class label $c$ are clustered tightly around the semantically similar prototypes $\vp_{c} \in \mP_c$. The cluster loss function is defined as
\begin{equation}
L_{\mathrm{Clst}} = - \frac{1}{N} \sum_{i = 1}^N 
        \frac{1}{|C|} \sum_{c \in C}
            \max_{\vp_{c} \in \mP_c}
                \max_{\vz_{c,i} \in \mZ_{c,i}}
                    s(\vp_c, \vz_{c,i}),
\end{equation}
where $s(\cdot, \cdot)$ is some similarity measure. The $L_{\mathrm{Clst}}$-Loss function increases compactness by increasing the similarity between prototypes $\vp_c$ and latent embeddings $\vz_c$ of class $c$ over all samples.

An additional separation loss increases the margin between different prototypes. The separation loss function is defined as
\begin{equation}
    L_{\mathrm{Sep}} =  \frac{1}{N} \sum_{i = 1}^N 
        \frac{1}{|C|} \sum_{c \in C}
            \max_{\vp_{\neg c} \notin \mP_c}
                \max_{\vz_{c,i} \in \mZ_{c,i}}
                    s(\vp_{\neg c}, \vz_{c,i}),
\end{equation}
The $L_{\mathrm{Sep}}$ function punishes high similarity values between a latent vector $\vz_c$ of class c and prototypes $\vp_{\neg c}$ not belonging to $c$, thereby separating the clusters in latent space.Please note, the original ProtoPNet implementation is optimized for minimizing the $L_2$ \textit{distance}, instead of maximizing the \textit{similarity} measure. Using the $L_2$ similarity would require tuning of the associated hyperparameters.

\subsection{Multi-Objective Loss Function}
To train a prototype learning network like HyperPGNet for downstream tasks like image classification, a multi-objective loss function is employed.
This multi-objective loss function is defined as
\begin{equation*}
    L = L_{\mathrm{CE}} + \lambda_{\mathrm{Clst}} L_{\mathrm{Clst}} + \lambda_{\mathrm{Sep}} L_{\mathrm{Sep}},
\end{equation*}
where $L_{\mathrm{CE}}$ is the cross-entropy loss over network predictions and ground truth image labels. Our experiments uses $\lambda_\mathrm{Clst} = 0.8$ and $\lambda_\mathrm{Sep} = 0.08$ as proposed by ProtoPNet \citep{Chen2019ThisLooksThat}. 

\input{tab/acc_table}

\subsection{Network Optimization}
ProtoPNet proposes a sophisticated network optimization scheme to achieve its high performance. 
Initially, for five warm-up epochs, only the convolution neck and prototype module are trained using Adam \cite{Kingma2015Adam} with learning rate \num{3e-3}, while the classification layer uses \num{1e-4}. During the warm-up phase, the pretrained image encoder remains frozen. 
After the warm-up phase, a new Adam optimizer trains the encoder with learning rate \num{1e-4}, while the neck and prototype module continue at \num{3e-3}. A scheduler reduces learning rates by \num{1e-1} every 5 epochs.

A preliminary Bayesian hyperparameter optimization using Weights and Biases \cite{wandb} yielded no significant improvements, suggesting the original ProtoPNet implementation is already highly optimized.

%% file: tab/acc_table.tex
\begin{table*}[t]
    \centering
    \footnotesize
    \setlength{\tabcolsep}{4pt}
    \caption{Mean Top-1 Test Accuracies (\%) CUB-200-2011.}
    \label{tab:all-results}
    \sisetup{
        separate-uncertainty=true,
        uncertainty-separator={\,\pm\,},
        detect-weight=true,
        detect-inline-weight=math
    }
    \resizebox{\textwidth}{!}{%
    \begin{tabular}{l S[table-format=2.1(1.1)] S[table-format=2.1(1.1)] @{\hskip 12pt} c @{\hskip 12pt} S[table-format=2.1(1.1)] S[table-format=2.1(1.1)] S[table-format=2.1(1.1)] @{\hskip 12pt} c @{\hskip 12pt} S[table-format=2.1, table-number-alignment=center] S[table-format=2.1, table-number-alignment=center]}
    \toprule
    & \multicolumn{2}{c}{\textbf{ProtoPNet Optim.}} && \multicolumn{3}{c}{\textbf{Simplified Optim.}} && \multicolumn{2}{c}{\textbf{Add. Datasets (R50)}} \\
    \cmidrule(lr){2-3} \cmidrule(lr){5-7} \cmidrule(lr){9-10}
    \textbf{Method} & {R50} & {D121} && {R50\textsuperscript{*}} & {D121\textsuperscript{*}} & {ViT\textsuperscript{*}} && {Cars\textsuperscript{*}} & {Flowers\textsuperscript{*}} \\ 
    \midrule
    Baseline        & 79.2 \pm 0.2 & 75.5 \pm 0.2 && 79.2 \pm 0.2 & 75.5 \pm 0.2 & 78.5 \pm 0.8 && 85.2 & 90.0 \\ 
    \cmidrule[0.5pt](lr){1-10}
    Euclidean (PPN)       & 73.9 \pm 0.4 & 76.4 \pm 1.2 && 61.4 \pm 1.4 & 57.8 \pm 3.0 & 23.4 \pm 0.4 && 75.4 & 70.7 \\
    Gaussian        & 6.4 \pm 0.3  & 75.7 \pm 0.2 && 61.1 \pm 0.2 & 45.9 \pm 5.1 & 4.2 \pm 1.3  && 73.0 & 74.8 \\
    \cmidrule[0.5pt](lr){1-10}
    Cosine          & \bfseries 78.5 \pm 0.2 & \bfseries 80.4 \pm 0.2 && 71.7 \pm 0.4 & 68.7 \pm 0.5 & 69.2 \pm 3.9 && \bfseries 79.6 & 85.5 \\
    HyperPG (Ours)  & 77.8 \pm 0.2 & 78.7 \pm 0.1 && \bfseries 74.3 \pm 0.6 & \bfseries 70.7 \pm 0.3 & \bfseries 72.9 \pm 0.1 && 79.0 & \bfseries 87.8 \\ 
    \bottomrule
    \multicolumn{10}{l}{\textsuperscript{*}Without warm-up epochs and training optimizations. R50: ResNet50, D121: DenseNet121} \\
    \end{tabular}
    }
\end{table*}

%% file: content/05_experiments.tex
\section{Experiments}
\autoref{tab:all-results} reports the mean top-1 accuracy and standard deviation for the tested models across three random seeds. Unlike prior work, we use only ImageNet weights, omitting model ensembles and pretrained iNaturalist weights. We first conduct image classification experiments on CUB-200-2011 \cite{Wah2011caltechucsdbirds} with the full architecture and training implementation as proposed by ProtoPNet \cite{Chen2019ThisLooksThat}, then with a simplified network optimization, and finally on two additional image classification datasets.
Implementation details such as data preprocessing or model implementation are provided in \autoref{app:implementation}.

\subsection{ProtoPNet Optimization}
We experiment on the CUB dataset using ResNet50 \cite{He2016Deepresiduallearning} and \mbox{DenseNet121 \cite{Huang2017Dense}} backbones pretrained on ImageNet with ProtoPNet's full optimization scheme \cite{Chen2019ThisLooksThat}. We also test a blackbox baseline without a prototype module. \autoref{tab:all-results} (ProtoPNet Optim.) presents the top-1 test accuracies on CUB-200-2011.

In this experiment, we evaluate different prototype formulations without adjusting hyperparameters. Cosine prototypes perform best, even outperforming the Dense121 blackbox baseline. While probabilistic formulations perform slightly worse than their point-based counterparts, Gaussian prototypes show high sensitivity to hyperparameters and backbone selection. They fail to learn meaningful representations with the ResNet50 backbone.

\subsection{Simplified Optimization}

To evaluate the robustness and practical applicability of different prototype 
formulations, we test them under a drastically simplified training scheme that 
removes ProtoPNet's multi-stage optimization, warm-up epochs, and learning rate 
scheduling in favor of a single AdamW optimizer  (learning rate \num{1e-4}, no scheduling) for the entire model. \autoref{tab:all-results} (Simplified Optim.) shows the top-1 accuracies on CUB-200-2011. The Euclidean prototype formulation shows heavily reduced performance, suggesting that the original ProtoPNet implementation was highly finetuned. While hyperspherical prototypes also perform worse, remain competitive with the fully-optimized Euclidean ProtoPNet.

\begin{figure}[h]
    \centering
    \includegraphics[width=0.8\linewidth]{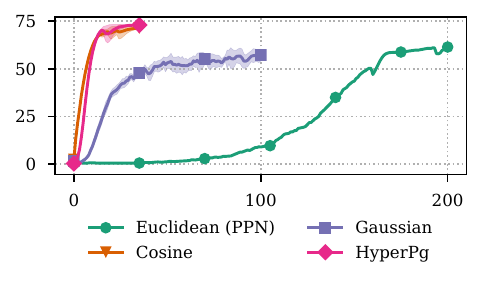}
    \caption{CUB-200-2011 Test Accuracy per Epoch with ResNet50 backbone and simplified optimization scheme. Mean and std over 3 random seeds.}
    \label{fig:acc-curves}
\end{figure}

\autoref{fig:acc-curves} illustrates the Top-1 Test Accuracy on CUB-200-2011 over training epochs. With the simplified training scheme, the hyperspherical prototype formulations converge much faster than $L_2$-based formulations. Cosine and HyperPg converge within 20 training epochs, whereas the Euclidean prototypes require over 200 epochs.

\subsection{Ablation: Transformer Backbone}
Using the simplified optimization scheme, we test a Vision Transformer backbone. Notably, both Euclidean and Gaussian prototypes fail to learn a latent structure from ViT-B16 features. We hypothesize that the Transformer's attention mechanism produces a near-hyperspherical latent space. This underscores the importance of matching prototype formulation to backbone architecture. For instance, language aligned models like CLIP \cite{radford21clip} use the cosine similarity for modality alignment. We anticipate that only hyperspherical prototypes can produce meaningful results when using CLIP as a feature extractor.

\subsection{Ablation: Datasets}
We apply our simplified training scheme to two additional datasets: Stanford Cars (Cars) with 196 car models \cite{Krause} and Oxford Flowers (Flowers) with 102 species \cite{Nilsback2008AutomatedFlower}. \autoref{tab:all-results} (Add. Datasets) shows top-1 test accuracies. The performance pattern among prototype formulations remains consistent across datasets, with Euclidean prototypes consistently underperforming.

%% file: content/06-xai.tex
\section{Interpretability Analysis}
We qualitatively analyze different prototype formulations with a ResNet50 backbone and the simplified optimization scheme. ProtoPNet originally proposed directly overlaying the prototype activation map (PAM). However, this visualization technique has several shortcomings, discussed in \autoref{app:xai}. As an alternative, we also present GradCAM \cite{GradCam} visualizations in \autoref{app:xai}.

\autoref{fig:xai-figs} shows PAMs for different prototype formulations on the same test image (left), with highest activations marked by yellow rectangles. Following the prototypical concepts approach \cite{ma2023this}, we present the two most highly activated training patches, as not all prototype formulations support a ``push'' onto training samples.

Interestingly, all prototype formulations activate in similar test image regions but differ in their associated training examples. The Euclidean prototype formulation seems to produce less visually consistent prototype associations than the other formulations using the simplified optimization scheme.

\begin{figure}[h!]
    \centering
    \includegraphics[width=1.0\linewidth]{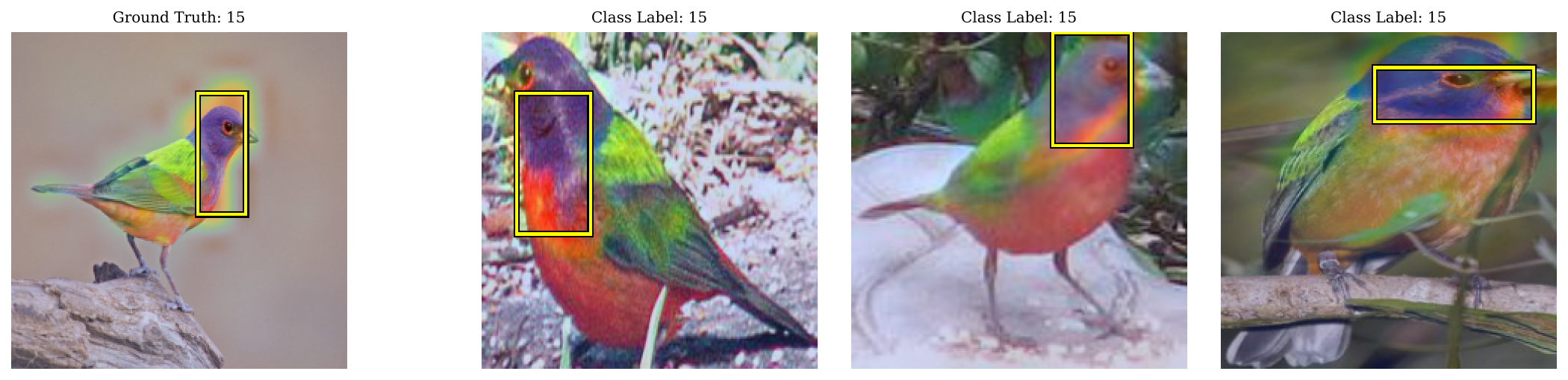}
    \caption{Cosine Protoypes with ViT. The overlayed heatmap show the distraction problem caused by the attention mechanism.}
    \label{fig:vit-problem}
\end{figure}

We observe no qualitative differences between visualizations produced by various prototype formulations using CNN-based backbones (see \autoref{app:xai}). While ViT backbone performs well with cosine prototypes, it exhibits the "distraction problem" (\autoref{fig:vit-problem}) first observed by ProtoPFormer \cite{Xue2024ProtoPFormer}. For an extensive overview of prototype examples, GradCAM activations, and additional backbones, refer to \autoref{app:xai}.

\input{figs/qual_xai}

%% file: figs/qual_xai.tex
\begin{figure}[h!]
    \centering
    \subfloat[\textbf{Euclidean} (R50)]{
        \includegraphics[width=1.0\linewidth]{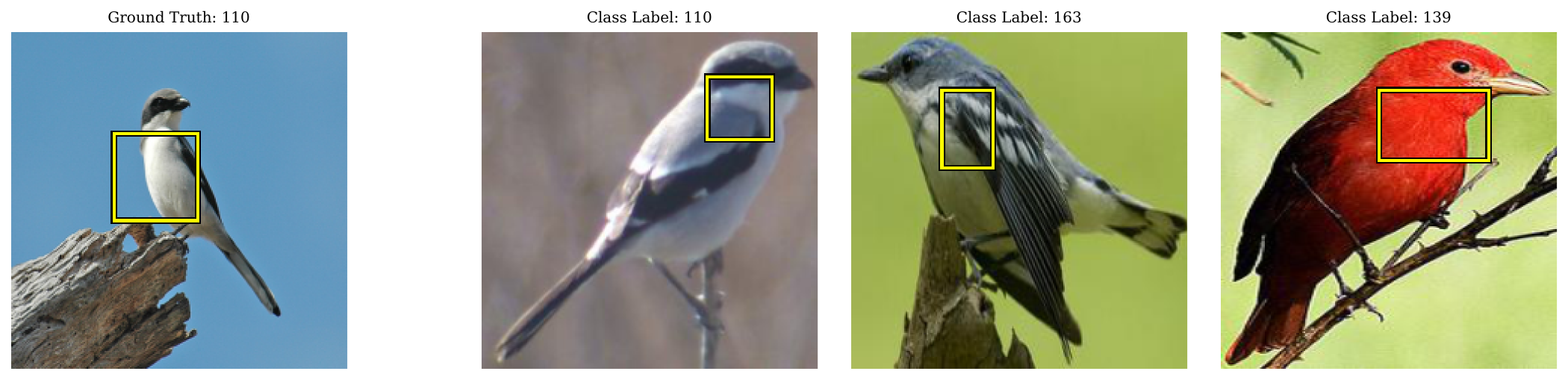}
    }\\
    \subfloat[\textbf{Gaussian} (R50)]{
        \includegraphics[width=1.0\linewidth]{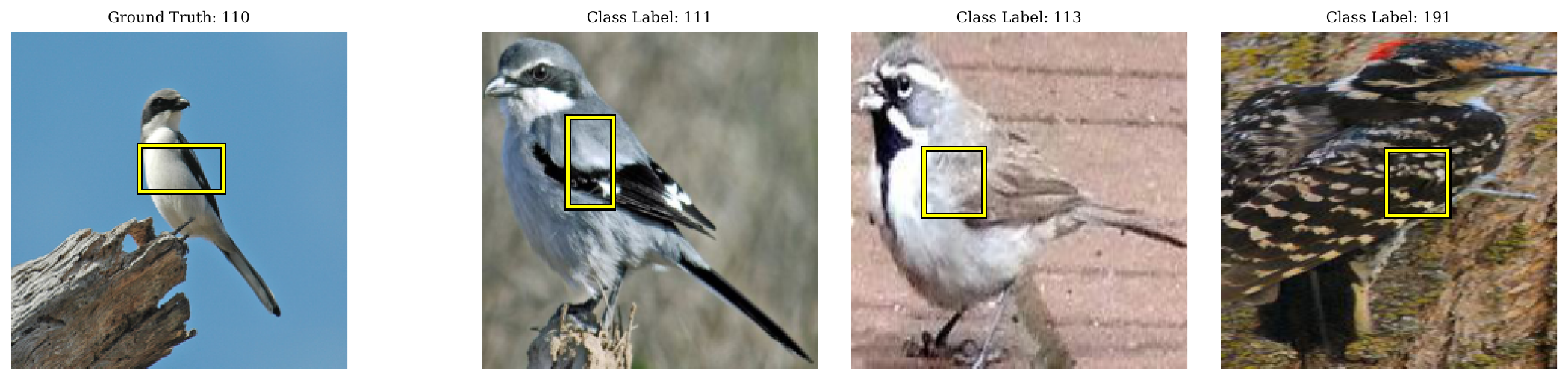}
    }\\
    \subfloat[\textbf{Cosine} (R50)]{
        \includegraphics[width=1.0\linewidth]{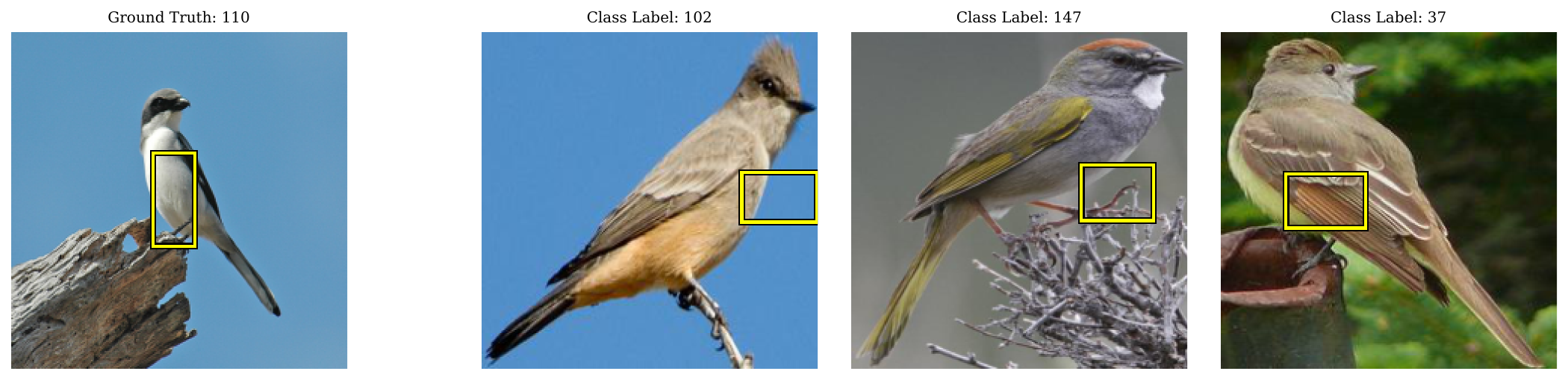}
    }\\
    \subfloat[\textbf{HyperPG} (R50)]{
        \includegraphics[width=1.0\linewidth]{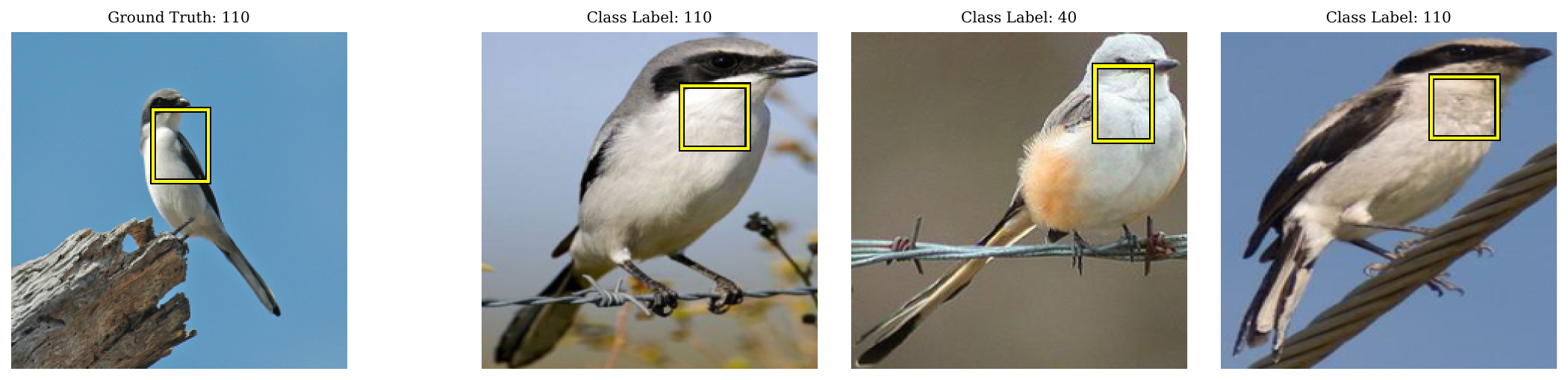}
    }\\  
    \caption{Left: Test Image. Right: Three closest Training Patches to the Prototype.}
    \label{fig:xai-figs}
\end{figure}

%% file: content/06_conclusion.tex
\section{Conclusion}
This work provides a comprehensive overview of different prototype formulations, with both probabilistic and point-based measurements, in Euclidean or hyperspherical latent spaces. We introduce HyperPG, a new prototype representation based on Gaussian distributions on the surface of a hypersphere. 

Our experiments demonstrate substantial differences in robustness across prototype formulations. Hyperspherical prototypes maintain competitive performance under drastically simplified training (single optimizer, no warm-up epochs or learning 
rate scheduling), whereas Euclidean prototypes show severe performance degradation under identical conditions. 
This indicates that the original ProtoPNet's strong performance depends heavily on extensive hyperparameter tuning, whereas hyperspherical formulations offer greater optimization stability. 
This has practical implications for deployment: hyperspherical prototypes provide improved performance with reduced sensitivity to training configuration, making prototype-based models more accessible for practitioners without extensive hyperparameter search resources.

Our results show the advantages of hyperspherical and probabilistic prototypes over the commonly used Euclidean formulation, with no loss to inherent interpretability. Future work could explore further refinements to probabilistic prototype formulations, including adaptive mechanisms for prototype selection or the integration of additional probabilistic models such as Mixture Models or Bayesian approaches like MGProto \cite{Wang2024MixtureGaussianDistributed}.

%% file: appendix/app-xai.tex
\clearpage
\section{Extended Interpretability Analysis}
\label{app:xai}

\begin{figure}[h]
    \centering
        \includegraphics[width=0.38\linewidth]{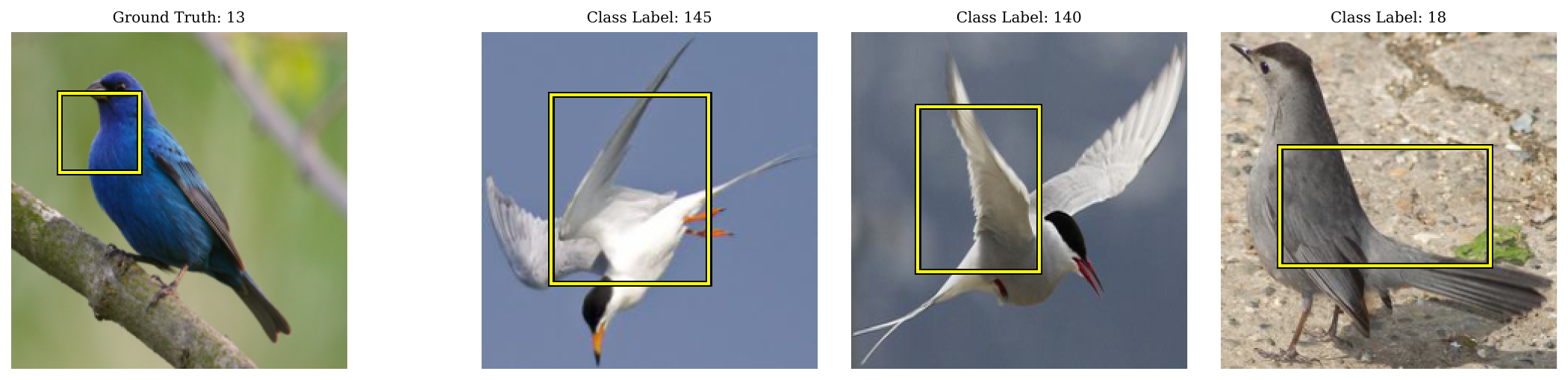}
        \hspace{0.5in}
        \includegraphics[width=0.38\linewidth]{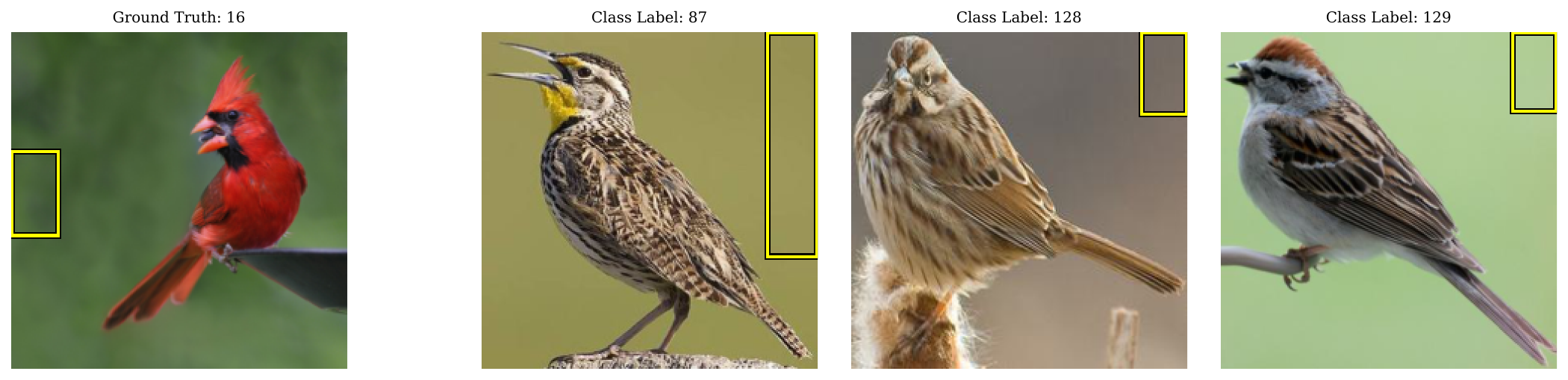}
        \vskip 0.1in
        \includegraphics[width=0.38\linewidth]{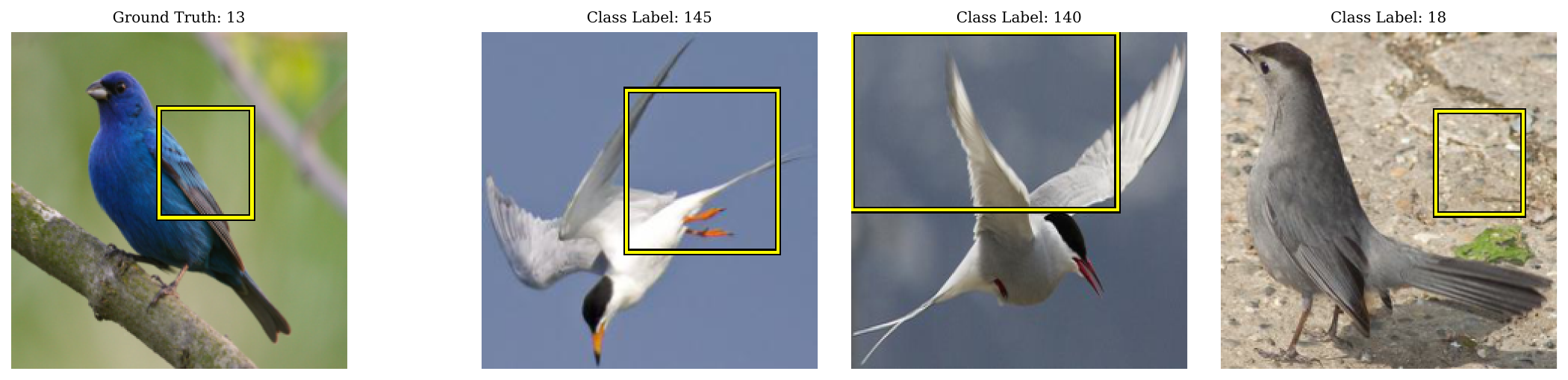}
        \hspace{0.5in}
        \includegraphics[width=0.38\linewidth]{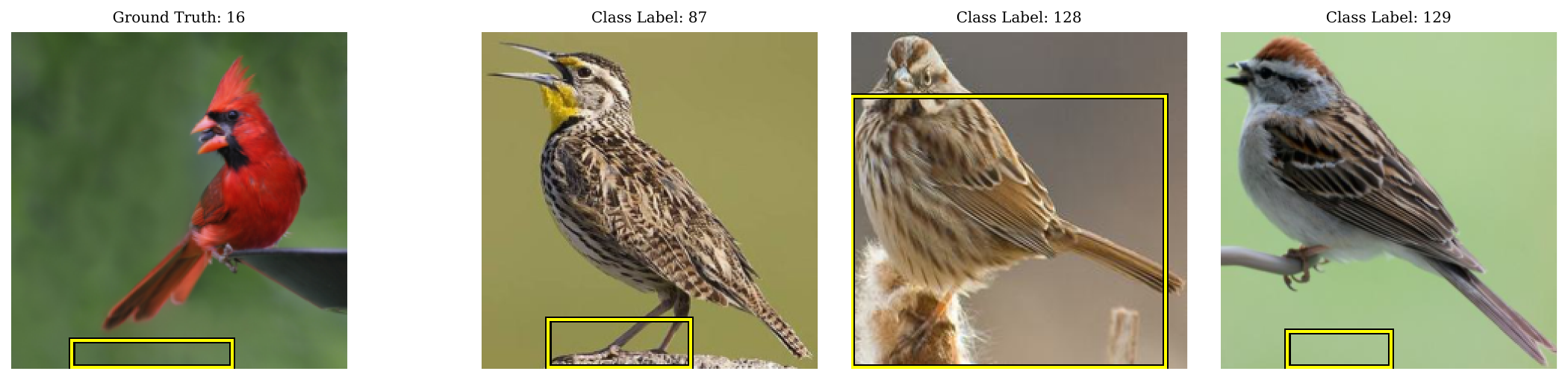}
    \caption{DenseNet121 \textbf{Euclidean}. Top: PAM. Bottom: GradCAM}
\end{figure}

\begin{figure}[h]
    \centering
        \includegraphics[width=0.38\linewidth]{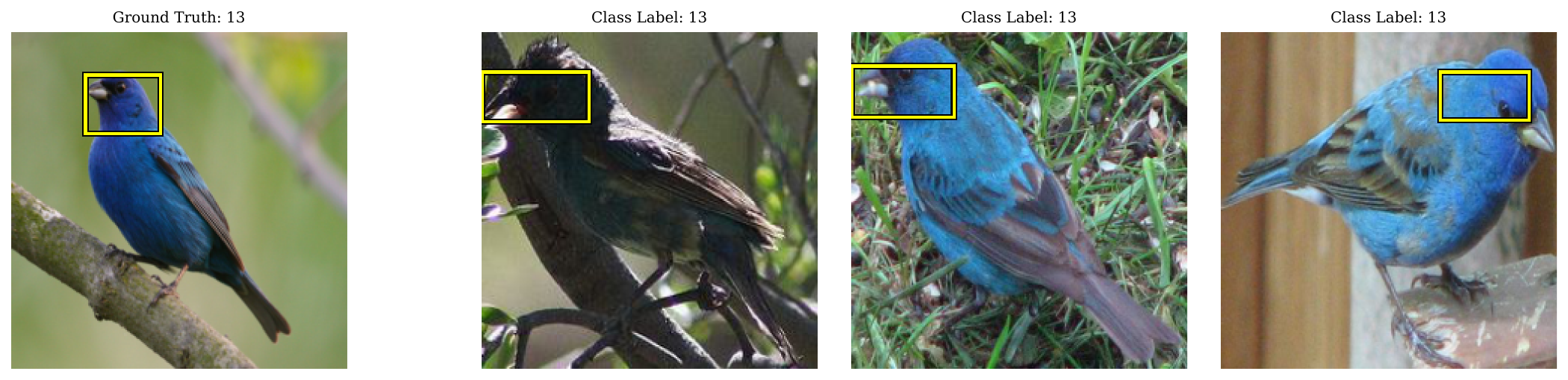}
        \hspace{0.5in}
        \includegraphics[width=0.38\linewidth]{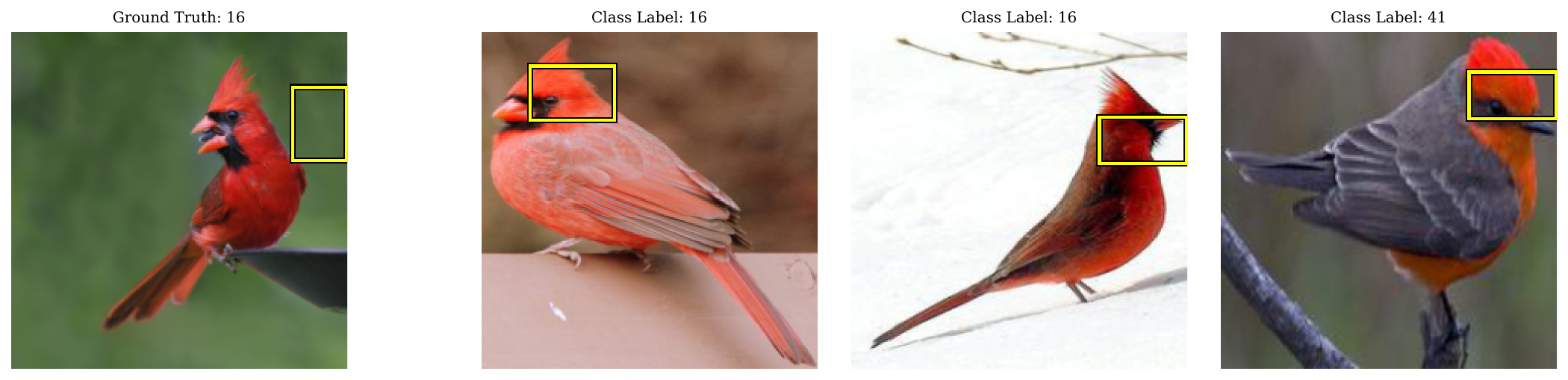}
        \vskip 0.1in
        \includegraphics[width=0.38\linewidth]{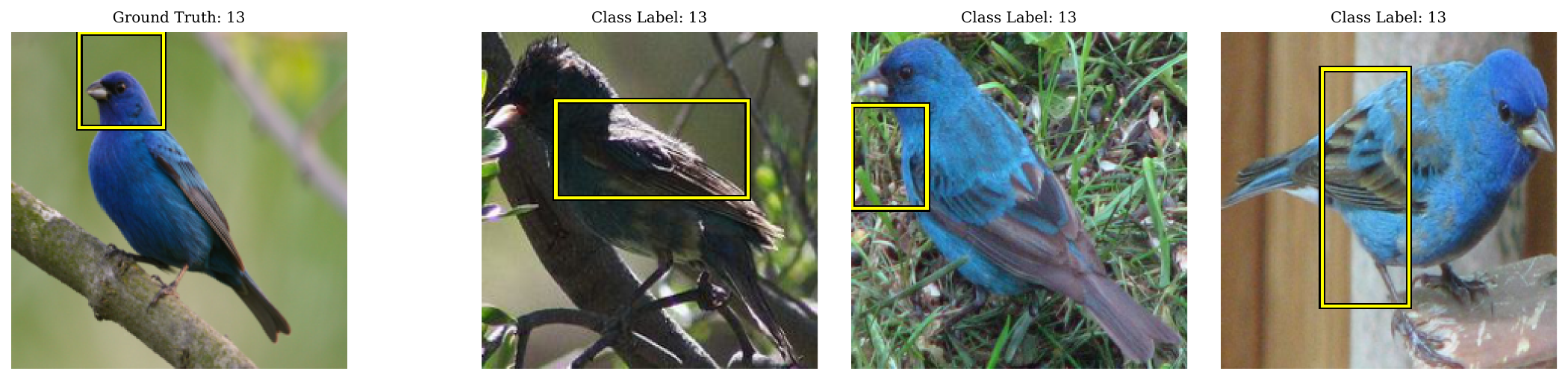}
        \hspace{0.5in}
        \includegraphics[width=0.38\linewidth]{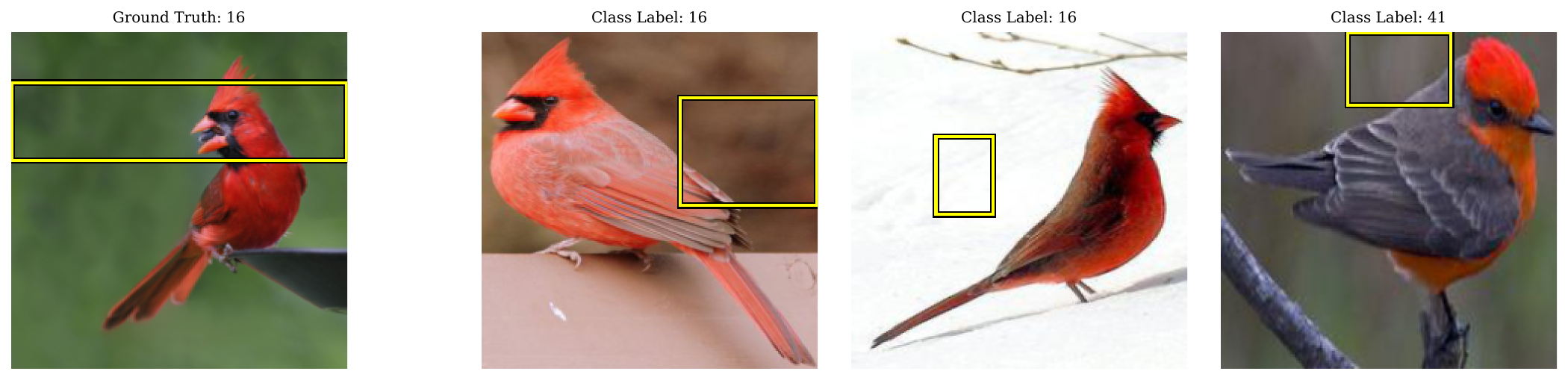}
    \caption{DenseNet121 \textbf{Gaussian}. Top: PAM. Bottom: GradCAM}
\end{figure}

\begin{figure}[h]
    \centering
        \includegraphics[width=0.38\linewidth]{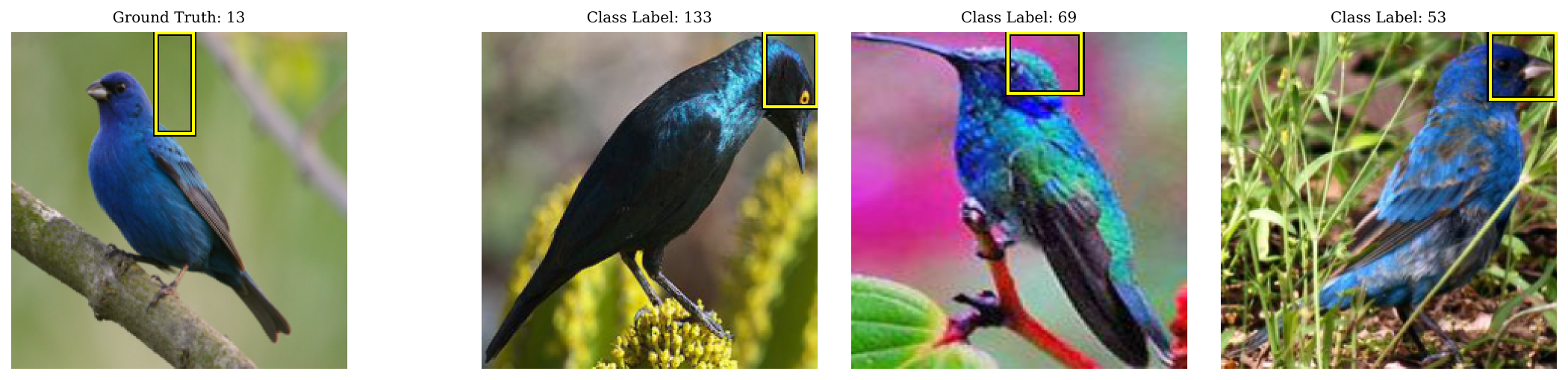}
        \hspace{0.5in}
        \includegraphics[width=0.38\linewidth]{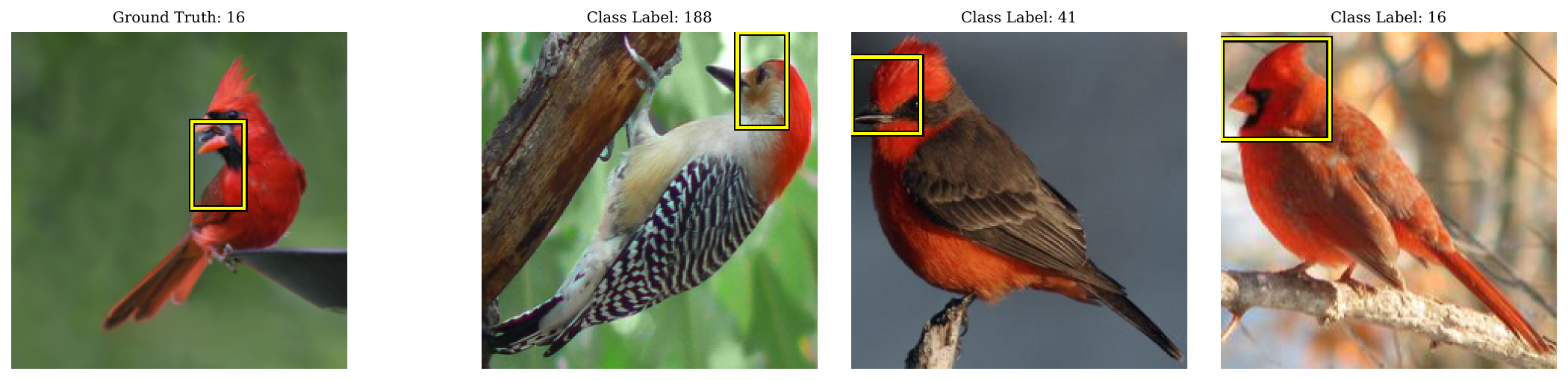}
        \vskip 0.1in
        \includegraphics[width=0.38\linewidth]{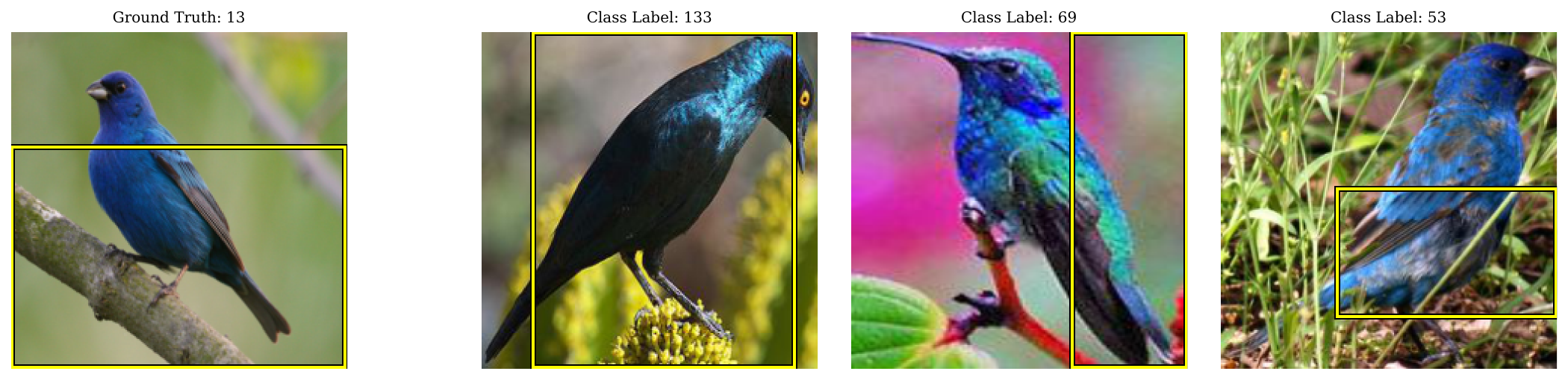}
        \hspace{0.5in}
        \includegraphics[width=0.38\linewidth]{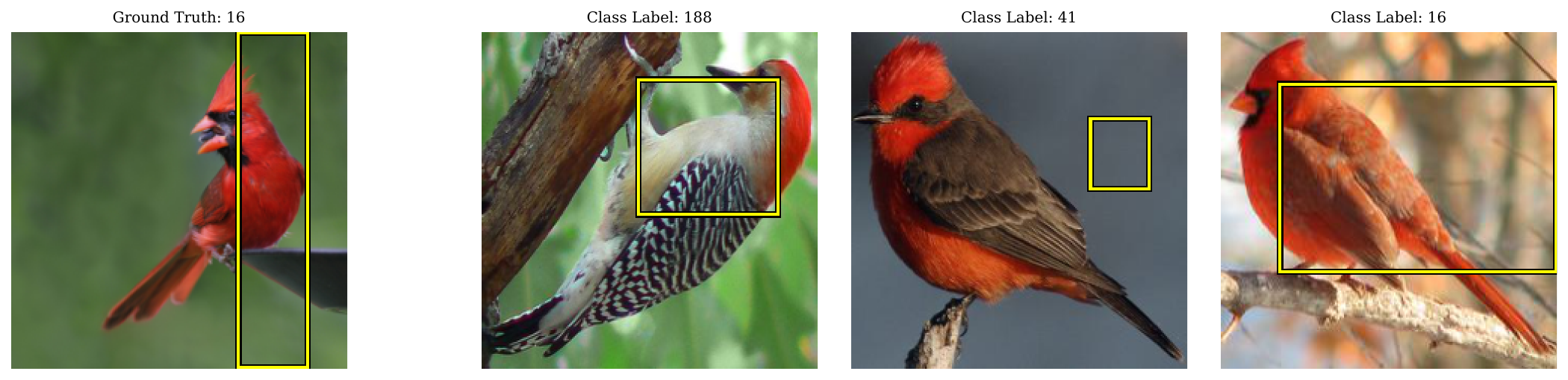}
    \caption{DenseNet121 \textbf{Cosine}. Top: PAM. Bottom: GradCAM}
\end{figure}

\begin{figure}[htb!]
    \centering
        \includegraphics[width=0.38\linewidth]{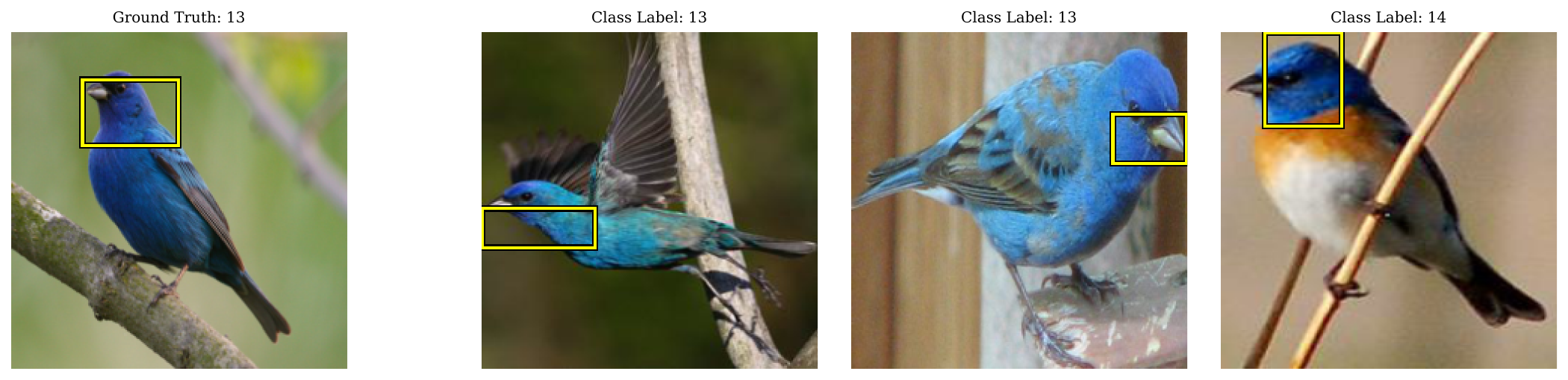}
        \hspace{0.5in}
        \includegraphics[width=0.38\linewidth]{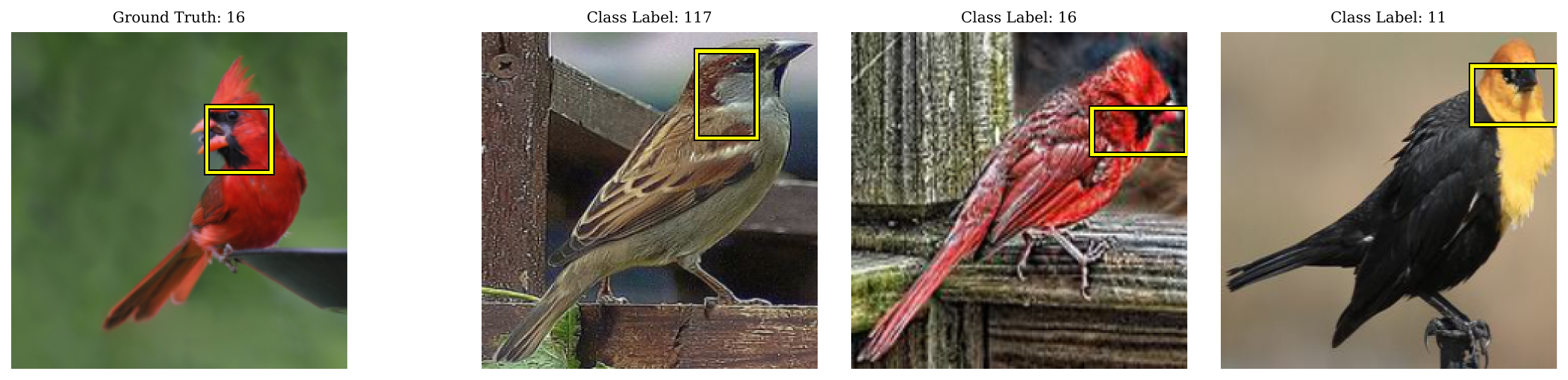}
        \vskip 0.1in
        \includegraphics[width=0.38\linewidth]{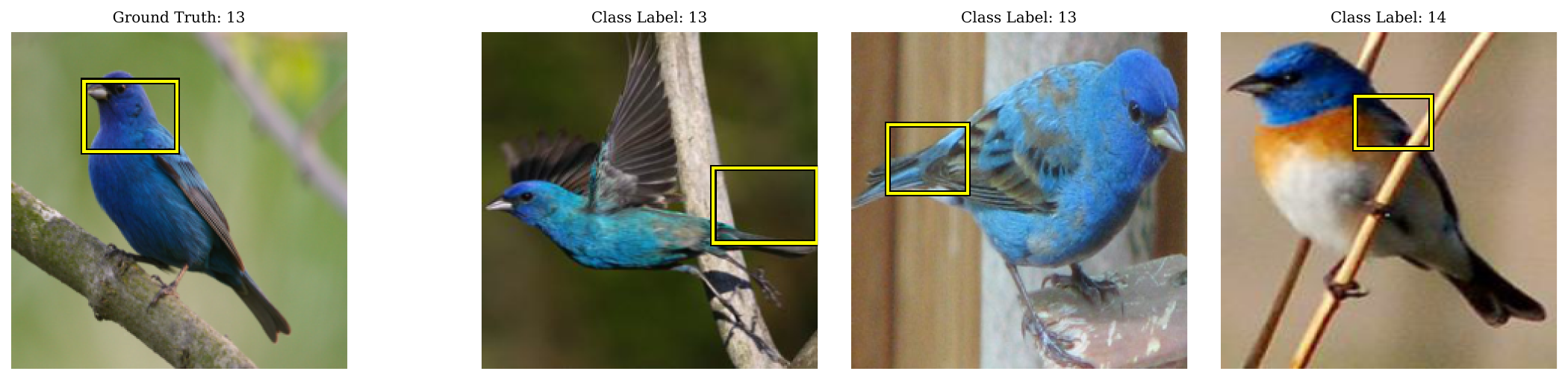}
        \hspace{0.5in}
        \includegraphics[width=0.38\linewidth]{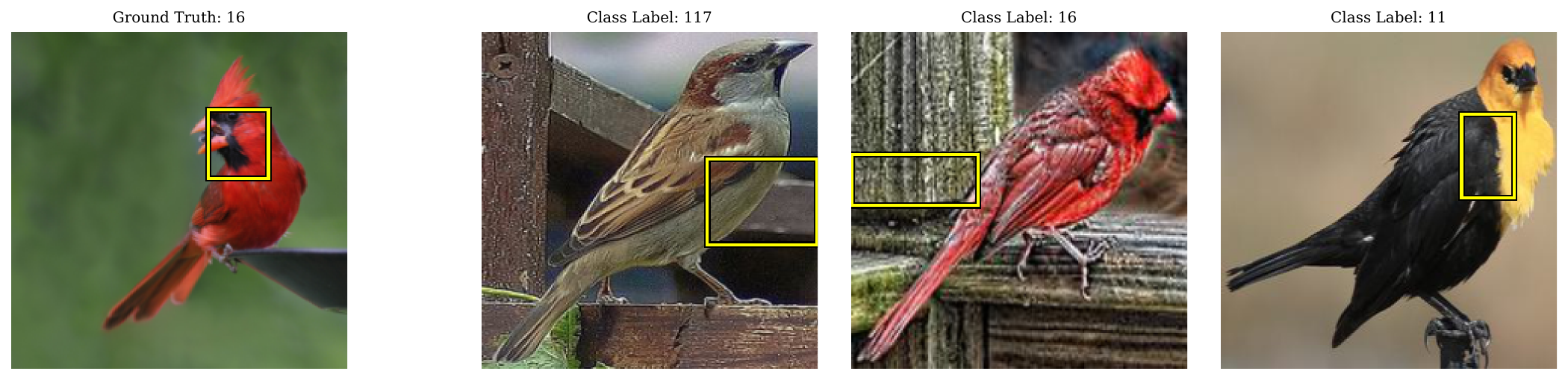}
    \caption{DenseNet121 \textbf{HyperPG}. Top: PAM. Bottom: GradCAM}
\end{figure}


\begin{figure}
    \centering
        \includegraphics[width=0.4\linewidth]{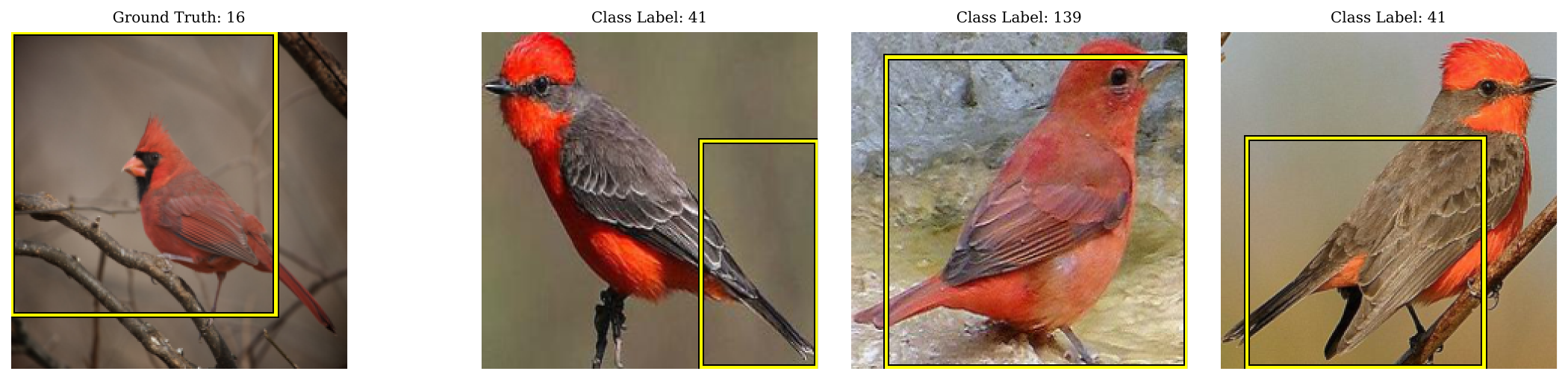}
        \hspace{0.5in}
        \vskip 0.15in
        \includegraphics[width=0.4\linewidth]{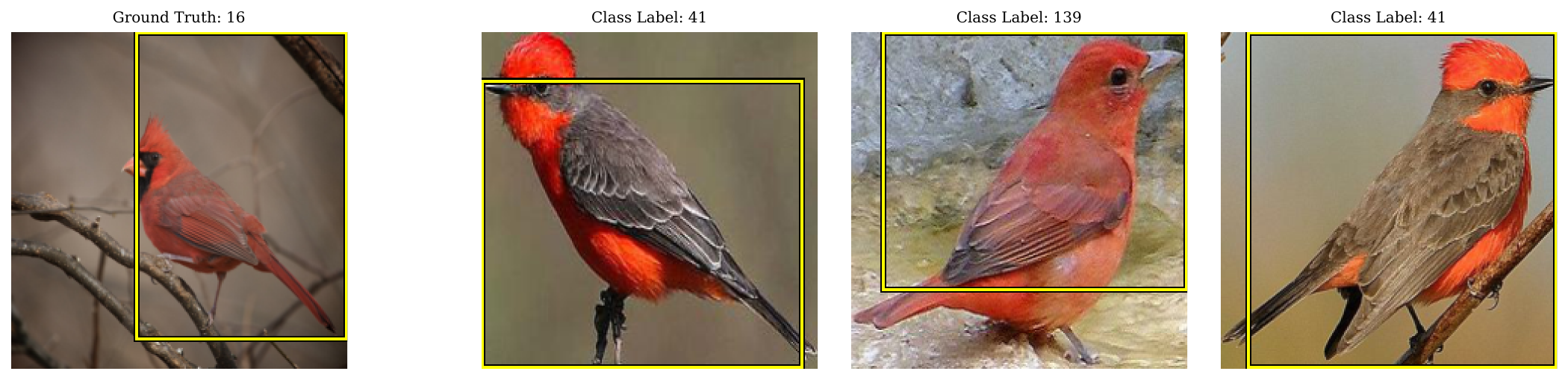}
        \hspace{0.5in}
    \caption{ViT-B16 \textbf{Euclidean}. Top: PAM. Bottom: GradCAM}
\end{figure}

\begin{figure}
    \centering
        \includegraphics[width=0.4\linewidth]{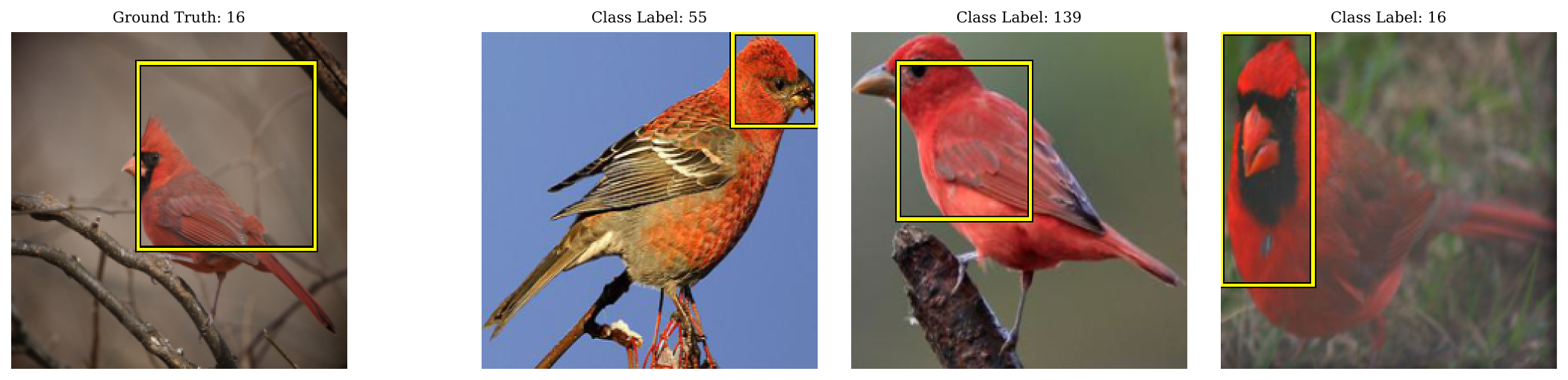}
        \hspace{0.5in}
        \vskip 0.15in
        \includegraphics[width=0.4\linewidth]{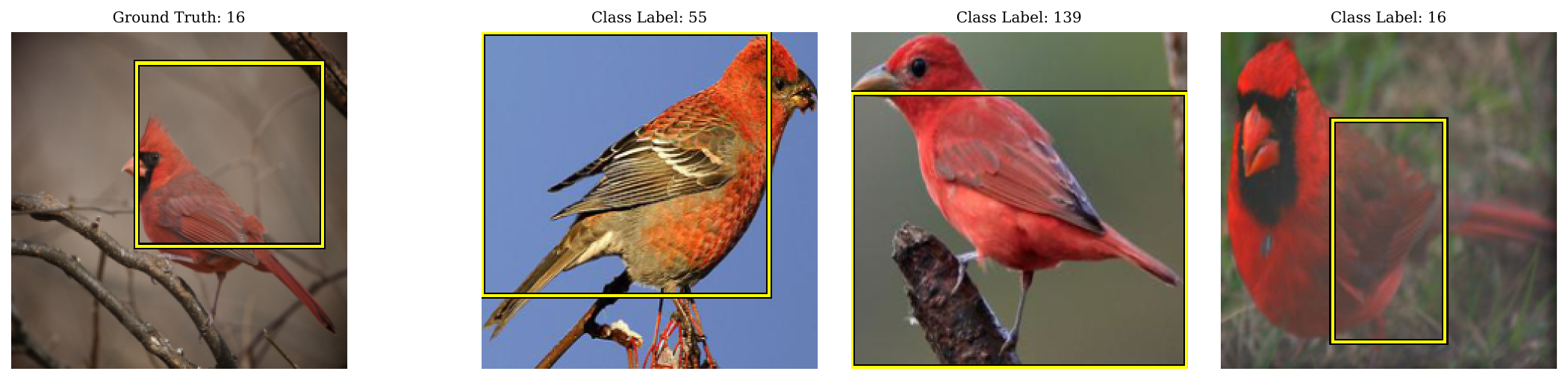}
        \hspace{0.5in}
    \caption{ViT-B16 \textbf{Gaussian}. Top: PAM. Bottom: GradCAM}
\end{figure}

\begin{figure}
    \centering
        \includegraphics[width=0.4\linewidth]{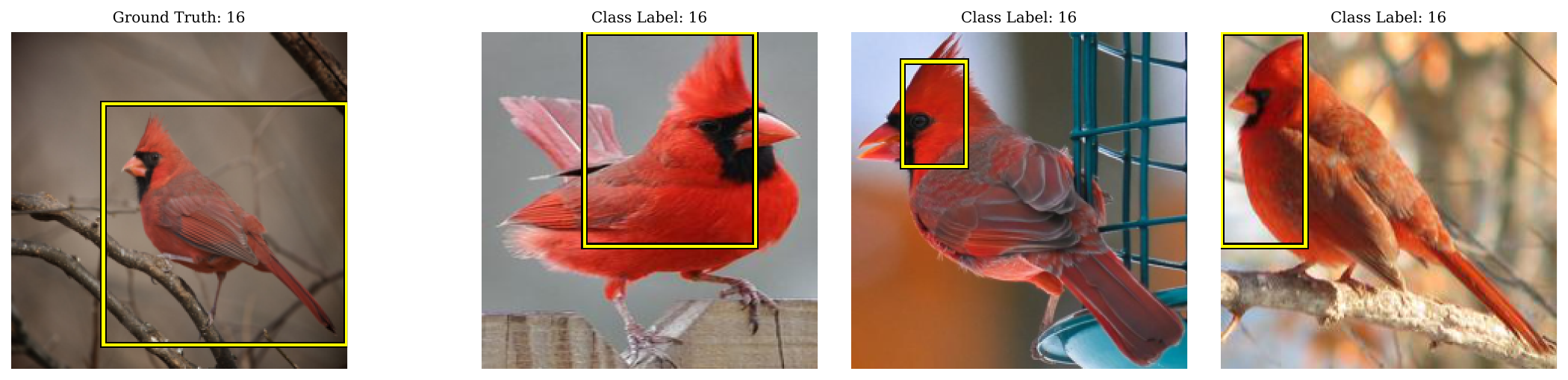}
        \hspace{0.5in}
        \vskip 0.15in
        \includegraphics[width=0.4\linewidth]{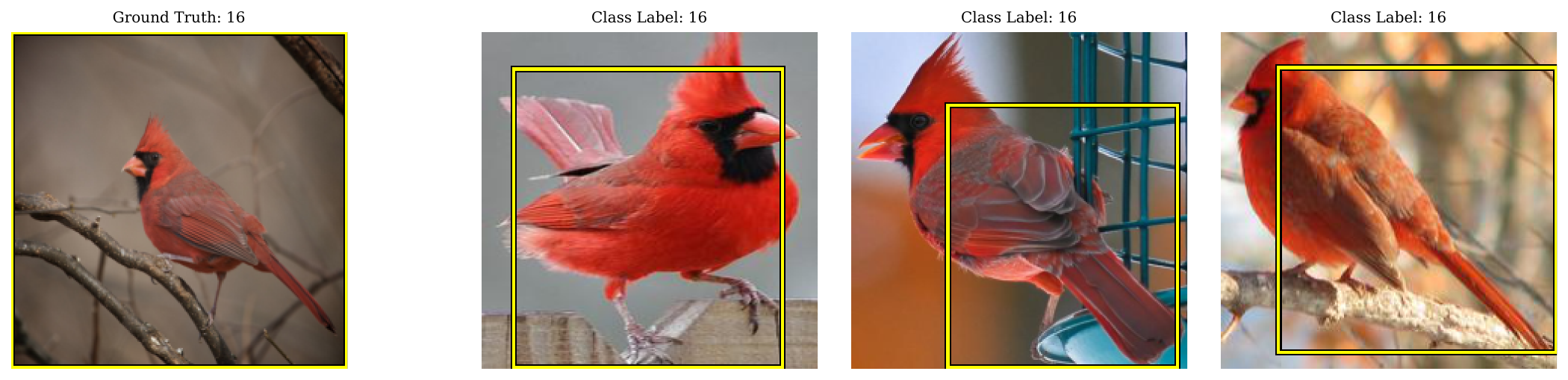}
        \hspace{0.5in}
    \caption{ViT-B16 \textbf{Cosine}. Top: PAM. Bottom: GradCAM}
\end{figure}

\begin{figure}
    \centering
        \includegraphics[width=0.4\linewidth]{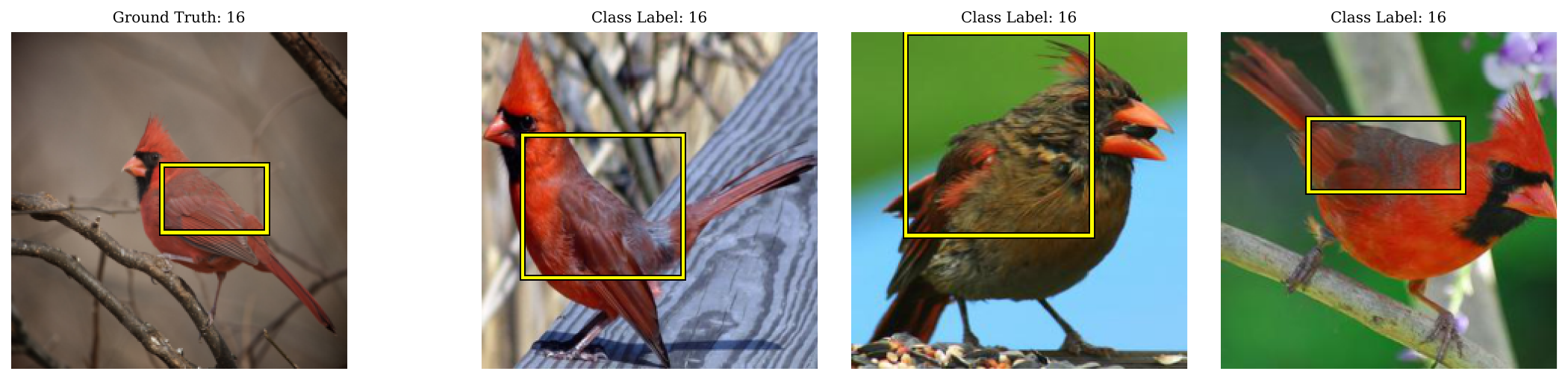}
        \hspace{0.5in}
        \vskip 0.15in
        \includegraphics[width=0.4\linewidth]{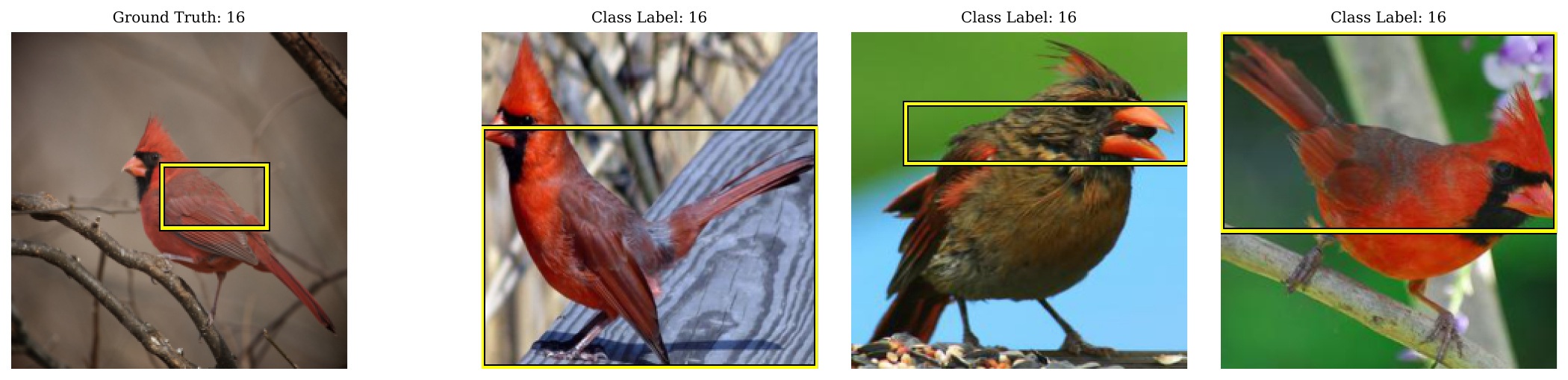}
        \hspace{0.5in}
    \caption{ViT-B16 \textbf{HyperPG}. Top: PAM. Bottom: GradCAM}
\end{figure}
\clearpage

\subsection{Potential Problems with Prototype Activation Maps}
One common visualization technique originally proposed by ProtoPNet is a a direct overlay of the prototype activation map (PAM). In this manner, the latent similarity map for each prototype is scaled back up to the original input resolution and overlayed over the image. A box is drawn around the 95th quantile to mark the highest activated image region. This technique has some obvious flaws: The latent similarity map is usually very low resolution, e.g., for ResNet50 $7 \times 7$, unless a segmentation model such as DeeplabV3 \cite{chen2017rethinkingatrousconvolutionsemantic} is used. Any perceived gradient in the overlay is likely an artifact from the upscaling operation. Additionally, there exist robustness problems with this visualization \cite{Sacha2024Interpretabilitybenchmarkevaluating}. By modifying parts of the background or adding noise to the image, the activation map can change, most likely due to the spatial pooling in the convolution pyramids.

%% file: appendix/app-implementation.tex
\section{Implementation Details}
\label{app:implementation}

\begin{table}[tbh]
\centering
\caption{Hyperparameters used in our experiments}
\label{tab:hyperparameters}
\begin{tabular}{lcc}
\toprule
\textbf{Hyperparameter} & \textbf{PPN} & \textbf{HyperPG} \\
\midrule
\multicolumn{3}{l}{\textit{Data and Model Configuration}} \\
Image size & 224 & 224\\
Number of prototypes per class & 10 & 10\\
Training batch size & 80 & 48\\
Testing batch size & 100 & 48\\
\midrule
\multicolumn{3}{l}{\textit{Training Schedule}} \\
Number of training epochs & 20 (offline augmentation $\times 30$) & 200 (online augmentation)\\
Number of warmup epochs & 5 & N.A.\\
\midrule
\multicolumn{3}{l}{\textit{Warm Optimizer Learning Rates}} \\
Add-on layers & $3 \times 10^{-3}$ & N.A.\\
Prototype vectors & $3 \times 10^{-3}$ & N.A.\\
\midrule
\multicolumn{3}{l}{\textit{Optimizer Learning Rates}} \\
Features & $1 \times 10^{-4}$ & $1 \times 10^{-4}$\\
Add-on layers & $3 \times 10^{-3}$ & $1 \times 10^{-4}$\\
Prototype vectors & $3 \times 10^{-3}$ & $1 \times 10^{-4}$\\
Joint learning rate step size & 5 & N.A.\\
\midrule
\multicolumn{3}{l}{\textit{Final Layer Training}} \\
Last layer optimizer learning rate & $1 \times 10^{-4}$ & $1 \times 10^{-4}$\\
\midrule
\multicolumn{3}{l}{\textit{Compute Resources}} \\
GPU Model & NVidia RTX 4090 & NVidia RTX 3070Ti \\
GPU VRAM & 24 GB & 8 GB \\
\bottomrule
\end{tabular}
\end{table}

Our experiments can be divided into two implementations: ``PPN'', built on the original ProtoPNet \cite{Chen2019ThisLooksThat} implementation published on Github\footnote{\url{https://github.com/cfchen-duke/ProtoPNet}} with a sophisticated network optimization scheme. And ``HyperPG'', our own implementation\footnote{\url{https://github.com/LiXiling/prob-proto}} built from scratch, originally designed for proposing and testing the HyperPG prototypes.

\subsection{Data Preprocessing}
\subsubsection{PPN: Data Preprocessing}
ProtoPNet uses an offline data augmentation process, resulting in 30 training images per class and epoch. The input images are cropped to the bounding box annotations. The images are augmented by applying Rotation XOR Skewing XOR Shearing, followed by a 50\% chance of left-right flipping.

\subsubsection{HyperPG: Data Preprocessing}
In contrast to prior work \citep[e.g.][]{Chen2019ThisLooksThat, Rymarczyk2020ProtopsharePrototypesharing, Ukai2023ThisLooksIt} the experiment used an online augmentation process, resulting in 30 training images per class and epoch. The input images were cropped to the bounding box annotations, then resized to a resolution of $224 \times 224$. The augmentations consisted of RandomPerspective, RandomHorizontalFlip and ColorJitter.

\subsection{Hyperparameters}
\label{app:hyperparam}
For both implementations we provide the full hyperparameters in \autoref{tab:hyperparameters}. PPN uses the exact hyperparameters as used by ProtoPNet according to the publically available Github implementation.

Our HyperPG implementation does not use a training scheme with warm-up epochs or learn-rate schedulings, making several hyperparameters not applicable, marked as N.A. The batchsizes are optimized for the different hardware configurations.

\subsection{Compute Resources}
\label{app:compute}
The PPN experiments were performed on a computing node with four NVidia RTX 4090 GPUs (24 GB VRAM each), with each model accessing one GPU. Please note, that the batch size could have been further increased on this hardware configuration. On CUB-200-2011 the training duration was roughly 20 minutes per epoch, with no meaningful difference between prototype formulations. 

The HyperPG experiments with simplified training and online data augmentation were performed on a desktop workstation with a single NVIDIA RTX 3070Ti GPU (8 GB VRAM). On CUB-200-2011 the training duration was roughly 40 seconds per epoch, with no meaningful difference between prototype formulations.

We estimate 160 GPU hours for running the PPN experiments plus 120 GPU hours for the HyperPG experiments. With an additional, generous 15\% overhead estimation for debugging, analyses, etc. we feel confident that this work took less than 320 GPU hours to develop.

\subsection{Model Implementation}
The experiments used various feature encoding backbones such as ResNet50 \citep{He2016Deepresiduallearning}, DenseNet121 \citep{Huang2017Dense} and ViT-B16 \cite{Dosovitskiy2020ImageIsWorth} with pretrained weights on ImageNet. All models use a single linear output layer as a classification head.

The prototypical networks follow the architecture proposed by ProtoPNet \citep{Chen2019ThisLooksThat}:
After the pretrained feature encoder, a bottleneck consisting of a Convolution Layer, ReLU activation, Convolution Layer and Sigmoid activation reduces the dimensions of the feature map to the prototype space. For example, the ResNet50 encoder produces a feature map of $H \times W \times D$ shape $7 \times 7 \times 2048$. The bottleneck produces a dimension reduction to shape $7 \times 7 \times 128$.

The prototype modules produce the prototype similarity scores, which are passed through a global max pooling, before being passed to the classification head.

%% file: content/05a_ablations.tex
\clearpage
\section{Additional Ablations}
\subsection{Ablation: Number of Prototypes}
\begin{figure}[h]
    \centering
    \includegraphics[width=0.4\linewidth]{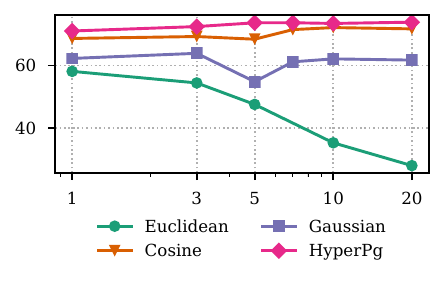}
    \caption{Ablation: Test Accuracy on CUB with different numbers of prototypes per class.}
    \label{fig:abl-numproto}
\end{figure}

We conduct an ablation study over the number of prototypes per class using the simplified training scheme and present the test accuracy scores in \autoref{fig:abl-numproto}. Notably, the predictive performance of Euclidean and Gaussian prototypes degrades when more prototypes are added. The prototype formulations with a hyperspherical structure, Cosine and HyperPG prototypes converge to similar performance levels regardless of the number of prototypes per class.

\subsection{Ablation: Number of Dimensions}
\begin{figure}[h]
    \centering
    \includegraphics[width=0.4\linewidth]{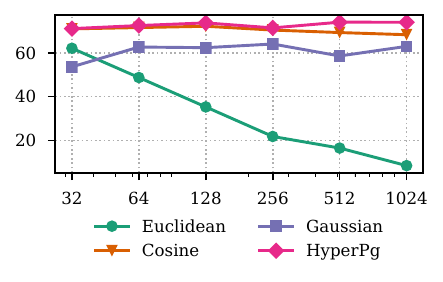}
    \caption{Ablation: Test Accuracy on CUB with different numbers of prototype dimensions.}
    \label{fig:abl-dim}
\end{figure}

We conduct an ablation study over the number of prototype dimensions using the simplified training scheme and present the test accuracy scores in \autoref{fig:abl-dim}. Hyperspherical prototype formulations such as Cosine and HyperPG are robust regarding the choice of prototype dimensions, although HyperPG performs slightly better with increased dimensionality. The Gaussian prototypes, like HyperPG prototypes, can adapt their variance $\sigma^2$ to deal with the increase in dimensionality. The performance of Euclidean prototypes suffers most with changes in dimensionality.

%% file: appendix/app-hyperpg-distr.tex
\clearpage
\section{Adapting HyperPG to other Probability Distributions}
\label{app:hyperpg-ext}
We define HyperPG prototypes $\vp^H = (\cvec{\alpha}, \mu, \sigma)$ as a Gaussian Distribution with mean $\mu$ and variance $\sigma^2$ of cosine similarities around an anchor vector $\alpha$. This idea of learning a distribution of cosine similarity values around an anchor $\cvec{\alpha}$ can be adapted to other distributions. This sections introduces some potential candidates. As early experiments on the CUB-200-2011 dataset showed no significant difference in performance, these sections are relegated to the appendix.

\subsection{Cauchy Distribution}
One theoretical disadvantage of the Gaussian distribution is the fast approach to zero, which is why a distribution with heavier tails such as the Cauchy distribution might be desirable.
The Cauchy distribution's PDF is defined as 
\begin{equation}
    \mathcal{C}(x; x_0, \gamma) = 
        \frac{1}{
            \pi \gamma \left(
                1 + \left(
                    \frac{x -x_0}{\gamma}
                    \right) ^2
                \right)
        },
\end{equation}
with median $x_0$ and average absolute deviation $\gamma$. The HyperPG prototypes with Cauchy are defined as accordingly as $\vp^{\mathrm{Cauchy}} = (\cvec{\alpha}, x_0, \gamma)$.

\begin{figure}[h]
    \centering
    \includegraphics[width=0.5\linewidth]{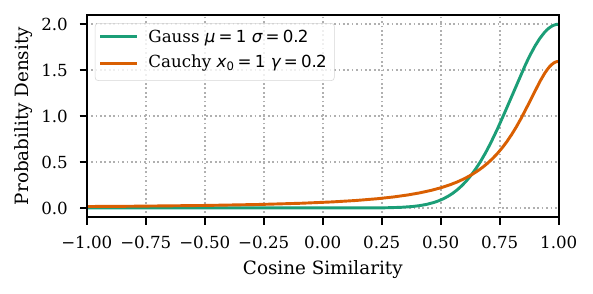}
    \caption{PDF for the Gaussian and Cauchy distribution of cosine similarity values. The Cauchy distribution has heavier tails, avoiding vanishing gradients issues.}
    \label{fig:gauss-cauchy}
\end{figure}

\autoref{fig:gauss-cauchy} illustrates the PDF of the Gaussian and Cauchy distributions with $\mu = x_0 = 1$ and $\sigma = \gamma = 0.2$, i.e., the main probability mass is aligned with the anchor $\cvec{\alpha}$. The Gaussian distributions PDF quickly approaches zero and stays near constant. This could potentially cause vanishing gradient issues during training. The heavier tails of the Cauchy distribution ensure that for virtually the entire value range of the cosine similarity, gradients could be propagated back through the model. However, experiments on CUB-200-2011 showed no significant performance difference between using HyperPG with the Gaussian or Cauchy distribution.

\subsection{Truncated Distributions}
The cosine similarity is defined only on the interval $[ -1, 1]$. This makes it attractive to also use truncated probability distributions, which are also only defined on this interval. The truncation imposes a limit on the range of the PDF, thereby limiting the influence of large values for the distribution's $\sigma$ or $\gamma$ parameter, respectively. The truncated Gaussian pdf $\mathcal{T}_{\mathrm{Gauss}}$ requires the cumulative probability function $\mathcal{G}$ and error function $f_{\mathrm{err}}$, and is defined as
\begin{align}
    f_{\mathrm{err}}(x) &=
        \frac{2}{\sqrt{\pi}}
            \int_0^x
                \exp\left(-z^2\right) dz
,\\
    \gN(x;\mu,\sigma) &= \frac{1}{\sqrt{2\pi\sigma^2}}
        \exp \left(
            - \frac{
                (x-\mu)^2
            }{
                2 \sigma^2
            }
        \right)
,\\
    \mathcal{G}(x, \mu, \sigma) &= \frac{1}{2}
        \left(
            1 + f_{\mathrm{err}}\left(
            \frac{x-\mu}{\sigma\sqrt{2}}
            \right)
        \right)
,\\
    \mathcal{T}_{\mathrm{Gauss}}(x,\mu,\sigma,a,b) &= 
        \frac{
        \gN\left(
            \scos; \mu, \sigma
        \right)
        }{
            \mathcal{G}(1, \mu, \sigma) - \mathcal{G}(-1, \mu, \sigma).            
        },
\end{align}
with lower bound $a$ and upper bound $b$, e.g., for the cosine similarity $a = -1$ and $b=1$. Similarly, the truncated Cauchy distribution can be applied, which is defined as
\begin{equation}
    \mathcal{T}_{\mathrm{Cauchy}}(x, x_0, \gamma, a, b) = 
        \frac{1}{\gamma}
        \left( 1 + \left( 
            \frac{x - x_0}{\sigma}
        \right)^2 \right)^{-1}
        \left(
        \arctan\left( \frac{b-x_0}{\gamma} \right)
            - \arctan\left( \frac{a-x_0}{\gamma} \right)
        \right)^{-1}.
\end{equation}

\begin{figure}
    \centering
    \includegraphics[width=0.5\linewidth]{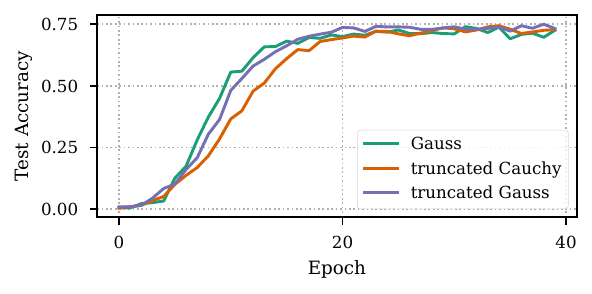}
    \caption{CUB-200-2011 Test Accuracy with HyperPG prototypes using different PDFs.}
    \label{fig:pdf-acc}
\end{figure}

\autoref{fig:pdf-acc} shows the test accuracy of three HyperPGNet models with Gaussian, truncated Gaussian and truncated Cauchy distribution on the CUB-200-2011 dataset. While difference in test performance and learning speed were minimal on the CUB-200-2011 dataset, further exploration is necessary, as other experiments showed that the concept-alignment on CUB-200-2011 dominates the learning process, lessening the influence of the prototypes.

\subsection{von Mises-Fisher Distribution}
\label{app:vmf}
The von Mises-Fisher distribution (vMF) is the analogue of the Gaussian distribution on the surface of a hypersphere \cite{Hillen2017Momentsvonmises}. The density function $f_d$ of the vMF distribution for a $D$-dimensional unit-length vector $\vv$ is defined as 
\begin{equation}
    f_d(\vv| \cvec{\alpha}, \kappa) = C_d(\kappa)\exp\left(\kappa \cvec{\alpha}^\top \vv\right),
\end{equation}
with mean vector $\cvec{\alpha}$, scalar concentration parameter $\kappa$ and normalization constant $C_d(\kappa)$. The normalization constant $C_d(\kappa)$ is a complex function and difficult to compute for higher dimensions, which is why, for example, Tensorflow\footnote{\href{https://www.tensorflow.org/probability/api_docs/python/tfp/distributions/VonMisesFisher}{Tensorflow API Documentation} -  Accessed 2026-01-07} only supports the vMF distribution for $D \leq 5$.

However, the vMF distribution is a viable similarity measure when using the unnormalized density function with $C_d(\kappa) = 1$. Working with unnormalized densities highlights the relationship between the normal distribution and the vMF distribution.

\begin{figure}[h]
    \centering
    \begin{minipage}[t]{0.2\linewidth}
        \centering
        \includegraphics[width=\textwidth]{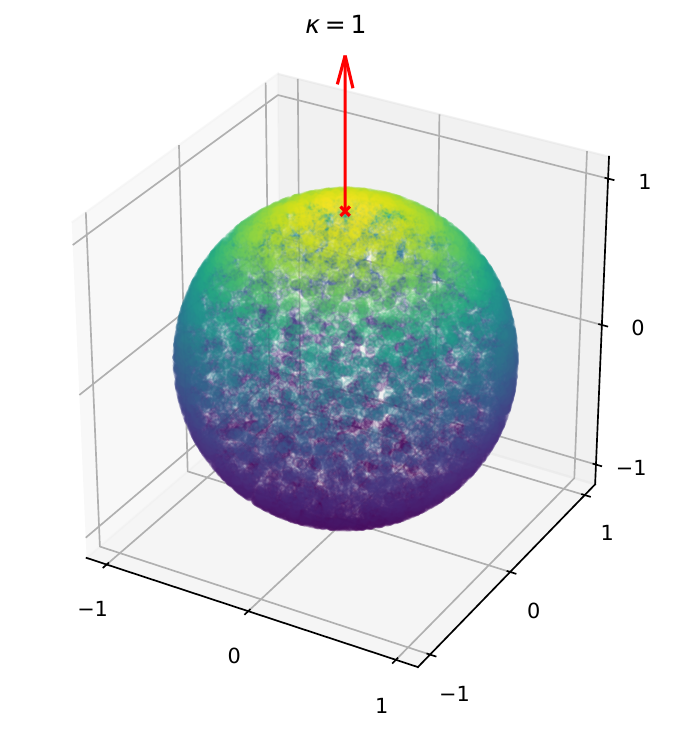}
    \end{minipage}
    \begin{minipage}[t]{0.2\linewidth}
        \centering
        \includegraphics[width=\textwidth]{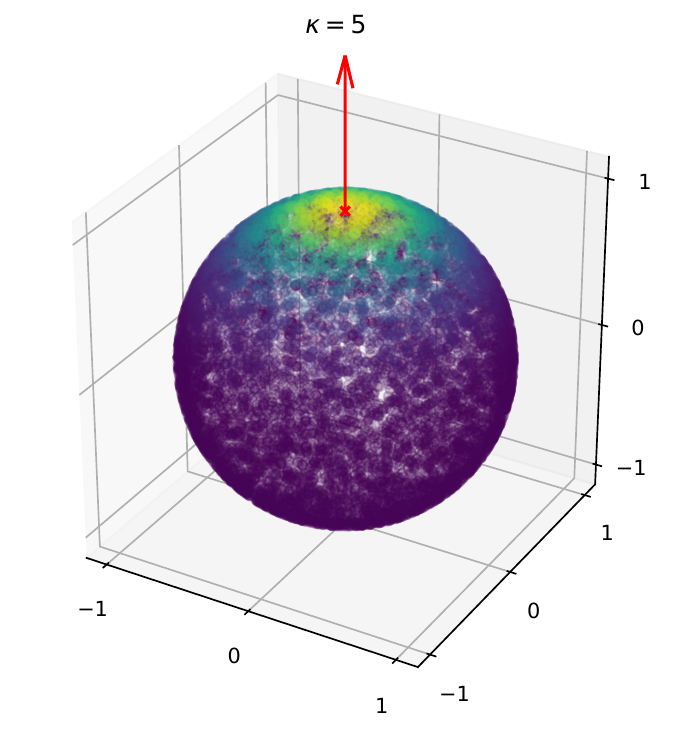}
    \end{minipage}
    \caption[]{Changing the concentration parameter $\kappa$ is akin to changing HyperPG's std $\sigma$.}
    \label{fig:vmf-dist}
\end{figure}

Let $\hat{G}$ be the unnormalized PDF of a multivariate Gaussian with normalized mean $\cvec{\alpha}$ and isotropic covariance $\cvec{\sigma}^2 = \kappa^{-1}\mI$, then it is proportional to the vMF distribution for normalized vectors $\vv$ with $|\vv|=1$, as shown by
\newcommand{\valpha}{\cvec{\alpha}}
\begin{align}
    \hat{G}(\vv | \valpha, \kappa)
        &= \exp \left(
            -\kappa \frac{(\vv -\valpha)^\top (\vv - \valpha)}{2}
            \right) \\
        &= \exp \left(
            -\kappa \frac{\vv^\top\vv + \valpha ^\top \valpha - 2 \vv^\top \valpha}{2}
            \right) \\
        &= \exp \left(
            -\kappa \frac{1 + 1 - 2 \vv^\top \valpha}{2}
            \right) \\
        &= \exp \left(
            -\kappa \frac{2 - 2 \vv^\top \valpha}{2}
            \right) \\
        &= \exp \left(
            -\kappa \frac{1 - \vv^\top \valpha}{1}
            \right) \\
        &= \exp \left(
            \kappa(\vv^\top \valpha - 1)
            \right) \label{eq:vmf-hyperpg}\\
        &= \exp(\kappa \vv^\top \valpha - \kappa) \\
        &= \exp(\kappa)^{-1} \exp(\kappa \vv^\top \valpha) \\
        &\sim \exp \left(
            \kappa\vv^\top \valpha
            \right).
\end{align}
\autoref{eq:vmf-hyperpg} also shows the relationship to the HyperPG similarity with an untruncated Gaussian distribution and prototype mean activation $\mu=1$.
\autoref{fig:vmf-dist} presents a simulation of the vMF distribution on a 3D sphere. While both the vMF distribution and HyperPG activation can produce a spherical, gaussian-like activation pattern on the surface of a hypersphere, the vMF distribution cannot produce the ring pattern shown in \autoref{fig:hyperpg-spheres}. The ring pattern produced by adapting HyperPG's mean similarity $\mu$ could be approximated by a mixture of vMF distributions.

\subsection{Fisher-Bingham Distribution}
As the vMF distribution is the equivalent of an isotropic Gaussian distribution on the surface of a hypersphere, the Fisher-Bingham (FB) distribution is the equivalent of a Gaussian with full covariance matrix. Similar to the vMF, the normalization constant is difficult to compute for higher dimensions, but the unnormalized density function remains feasible.

For a $D$ dimensional space, the FB distribution is by a $D \times D$ matrix $\mA$ of orthogonal vectors $(\valpha_1, \valpha_2, \ldots, \valpha_D)$, concentration parameter $\kappa$ and ellipticity factors $[\beta]_{2:D}$ where $\sum_{j=2}^D \beta_j = 1$ and $0 \leq 2 |\beta_j| < \kappa$. The FB's unnormalized PDF is defined as
\begin{equation}
    b(\vv| \mA, \kappa, \beta) =  \exp \left( \kappa \valpha_1^\top\vv + \sum_{j=2}^D \beta_j \left(\valpha_j^\top \vv\right)^2 \right).
\end{equation}

\begin{figure}
    \centering
    \includegraphics[width=0.25\linewidth]{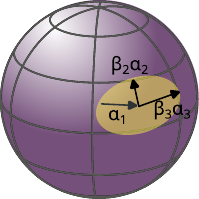}
    \caption{Illustration of the Fisher-Bingham Distribution in $D=3$}
    \label{fig:fb-distr}
\end{figure}

The FB distribution's main advantage is the elliptic form of the distribution on the surface of the hypersphere, offering higher adaptability than the other formulations (see \autoref{fig:fb-distr}. However, the parameter count and constraints are higher.

\subsection{Mixture Models}
HyperPG's probabilistic nature lends itself to a mixture formulation. Let the definition of a HyperPG Mixture Prototype be $\vp^M = (\cvec{\alpha}, \mu, \sigma, \pi)$ with additionally learned mixture weight $\pi$. Further, let's define the probability of a latent vector $\vz$ belonging to a Gaussian HyperPG prototype $\vp$ as
\begin{align}
    \phi (\vz | \vp) = s_{\mathrm{HyperPG}}(\vz|\vp).
\end{align}

Then the probability of $\vz$ belonging to class $c$ can be expressed through the mixture of all prototypes $\vp_c \in \mP_c$ of class $c$, i.e.,
\begin{equation}
    \phi (\vz | c) = \sum_{\vp_c \in \mP_c} \pi(\vp_c) \phi(\vz | \vp_c).
\end{equation}
First experiments with mixture of HyperPG prototypes did not show any improvement over the standard formulation. However, this might change with other datasets.

%% file: appendix/app-scaleddot.tex
\clearpage
\section{Scaled Dot-Product Prototype}
\label{app:scaleddot}
This section does not reiterate the full attention mechanism. We recommend to the interested reader to refer to the original work \cite{Vaswani2017Attention}. Rather, we will take a high-level view of the main equation and explain how we apply it to a prototype formulation.

\citet{Vaswani2017Attention} define the scaled dot-product attention for some input matrices query $\mQ$, key $\mK$ and value $\mV$ as
\begin{equation}
    \mathrm{Attention}(\mQ,\mK,\mV) = \mathrm{softmax}\left(\frac{\mQ\mK^\top}{\sqrt{d_k}} \right)\mV
\end{equation}
with $d_k$ being the number of dimensions for the key matrix $\mK$.

\begin{figure}[h]
    \centering
    \includegraphics[width=0.5\linewidth]{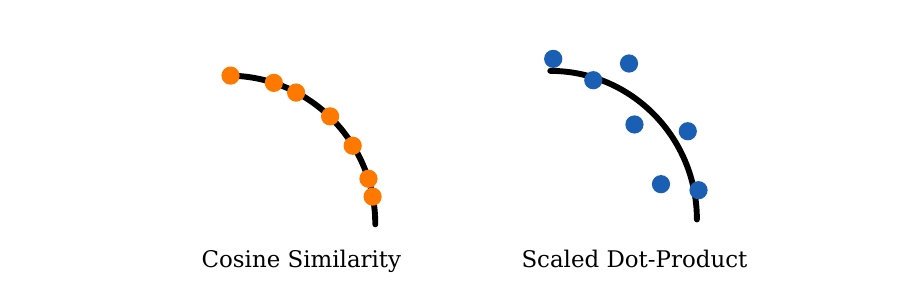}
    \caption{Illustration of Cosine Similarity and Scaled Dot-Product}
    \label{fig:ill-scaled-dot}
\end{figure}

We can use the scaled dot-product to define a new prototype similarity function
\begin{equation}
    s_\mathrm{sdot}(\vz|\vp) = \frac{\vz^\top \vp}{\sqrt{D}}
\end{equation}
for a latent patch $\vz$ and prototype vector $\vp$ with dimensionality $D$. In contrast to the cosine similarity, the vectors are not normalized to unit length, which preserves more information. However, dividing by the square root of the number of dimensions $D$ acts as a heuristic regarding vector length. Intuitively we can think about it as instead of a hard projection onto the surface of the unit-radius hypersphere, all points are moved \textit{towards} the surface. \autoref{fig:ill-scaled-dot} illustrates this scaling. In this manner, the angle between vectors becomes a meaningful measurement, while also retaining some information encoded by the vector length.

%% file: appendix/app-hyperpg-params.tex
\clearpage
\section{Parameter Distribution of HyperPG}
To evaluate, if HyperPG actually uses the probabilstic components, we look at the learned mean $\mu$ and standard deviation $\sigma$ values after training on CUB200-2011.

\begin{figure}[h]
    \centering
    \includegraphics[width=0.6\linewidth]{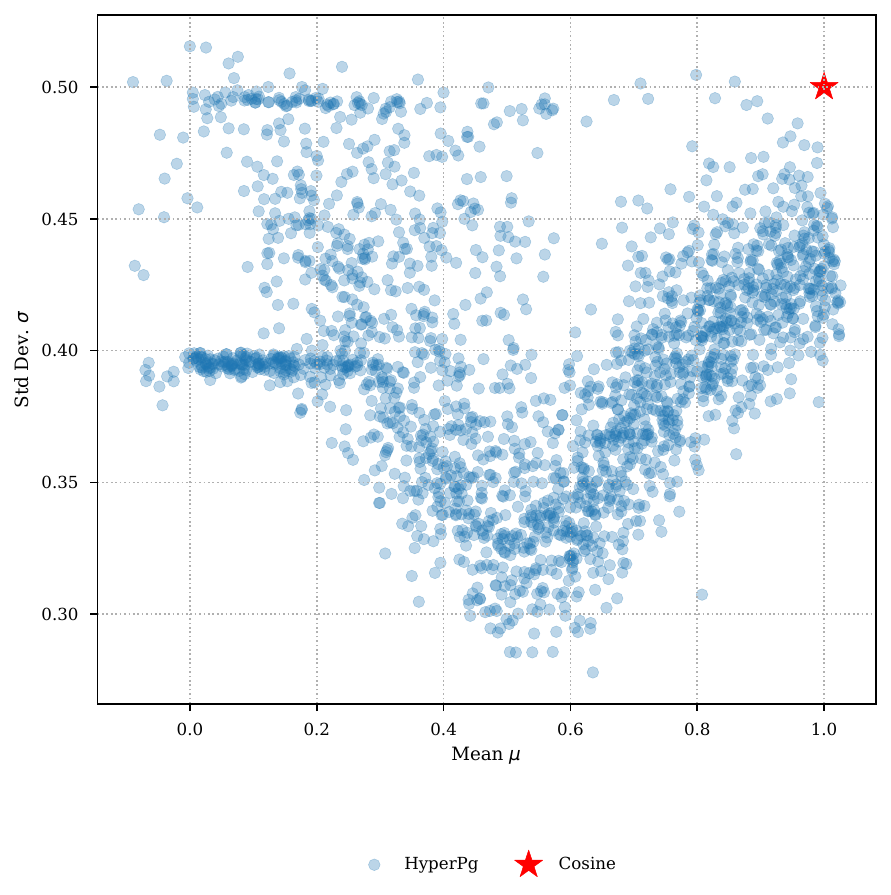}
    \caption{HyperPG learned mean and sigma values.}
    \label{fig:app-mean-sigma}
\end{figure}

\autoref{fig:app-mean-sigma} plots the mean and sigma values. The red star marks the theoretical solution if it would learn something similar to the pure Cosine prototypes. The pattern demonstrates, that HyperPG actually learns different parameters for the Gaussian components of each prototype. Interestingly, two modes at $\sigma=0.5$ and $\sigma=0.4$ appear with a distinct line shape, which warrant future exploration.

\clearpage
\section{Asset Licenses}
\label{app:license}
Our work relies on the publicly available Github Code of ProtoPNet \cite{Chen2019ThisLooksThat} which is released under the MIT license.

Our experiments use the following datasets:
\begin{itemize}
    \item CUB-200-2011 \cite{Wah2011caltechucsdbirds} by Caltech Vision Lab. The use is restricted to non-commercial research and educational purposes.
    \item Oxford Flowers \cite{Nilsback2008AutomatedFlower} from the Visual Geometry Group at University of Oxford. No license is detailed.
    \item  Stanford Cars \cite{Krause} as published on Kaggle. No license is detailed.
\end{itemize}

%% file: main.bib
@Article{Zhou2022RethinkingSemanticSegmentation,
  author  = {Zhou, Tianfei and Yang, Yi and Konukoğlu, Ender and Goo, Luc Van},
  journal = {Computer Vision and Pattern Recognition},
  title   = {Rethinking Semantic Segmentation: A Prototype View},
  year    = {2022},
  doi     = {10.1109/cvpr52688.2022.00261},
  mag_id  = {4313023122},
}

@Article{Ukai2023ThisLooksIt,
  author  = {Ukai, Y. and Hirakawa, Tsubasa and Yamashita, Takayoshi and Fujiyoshi, H.},
  journal = {International Conference on Learning Representations},
  title   = {This Looks like It Rather Than That: Protoknn for Similarity-Based Classifiers},
  year    = {2023},
}

@Article{Chen2019ThisLooksThat,
  author  = {Chen, Chaofan and Li, Oscar and Tao, Daniel and Barnett, Alina and Rudin, Cynthia and Su, Jonathan K.},
  journal = {Neural Information Processing Systems},
  title   = {This Looks like That: Deep Learning for Interpretable Image Recognition},
  year    = {2019},
  mag_id  = {2971048680},
  ranking = {rank1},
}

@InProceedings{Wang2024MCPNetInterpretableClassifier,
  author    = {Wang, Bor-Shiun and Wang, Chien-Yi and Chiu, Wei-Chen},
  booktitle = {Proceedings of the IEEE/CVF Conference on Computer Vision and Pattern Recognition},
  title     = {Mcpnet: An Interpretable Classifier Via Multi-Level Concept Prototypes},
  year      = {2024},
  pages     = {10885--10894},
}

@Article{Moradinasab2024ProtoGMMMultiprototype,
  author  = {Moradinasab, Nazanin and Shankman, Laura S. and Deaton, Rebecca A. and Owens, Gary K. and Brown, Donald E.},
  journal = {arXiv preprint arXiv:2406.19225},
  title   = {Protogmm: Multi-Prototype Gaussian-Mixture-Based Domain Adaptation Model for Semantic Segmentation},
  year    = {2024},
}

@Article{Nauta2023PIPNetPatch,
  author  = {Nauta, Meike and Schlötterer, Jörg and van Keulen, Maurice and Seifert, Christin},
  journal = {Computer Vision and Pattern Recognition},
  title   = {Pip-Net: Patch-Based Intuitive Prototypes for Interpretable Image Classification},
  year    = {2023},
  doi     = {10.1109/cvpr52729.2023.00269},
  mag_id  = {4386065251},
}

@Article{Donnelly2021DeformableProtoPNetInterpretable,
  author  = {Donnelly, Jonathan and Barnett, A. and Chen, Chaofan},
  journal = {Computer Vision and Pattern Recognition},
  title   = {Deformable Protopnet: An Interpretable Image Classifier Using Deformable Prototypes},
  year    = {2021},
  doi     = {10.1109/cvpr52688.2022.01002},
}

@Article{Rymarczyk2020ProtopsharePrototypesharing,
  author  = {Rymarczyk, Dawid and Struski, {\L}ukasz and Tabor, Jacek and Zieli{\'n}ski, Bartosz},
  journal = {arXiv preprint arXiv:2011.14340},
  title   = {Protopshare: Prototype Sharing for Interpretable Image Classification and Similarity Discovery},
  year    = {2020},
}

@InCollection{Lundberg2017UnifiedApproachInterpreting,
  author    = {Lundberg, Scott M. and Lee, Su-In},
  booktitle = {Advances in Neural Information Processing Systems 30},
  publisher = {Curran Associates, Inc.},
  title     = {A Unified Approach to Interpreting Model Predictions},
  year      = {2017},
  editor    = {I. Guyon and U. V. Luxburg and S. Bengio and H. Wallach and R. Fergus and S. Vishwanathan and R. Garnett},
  pages     = {4765--4774},
}

@Book{Molnar2020Interpretablemachinelearning,
  author    = {Molnar, Christoph},
  publisher = {Lulu. com},
  title     = {Interpretable Machine Learning},
  year      = {2020},
}

@InProceedings{Sacha2023ProtosegInterpretablesemantic,
  author    = {Sacha, Miko{\l}aj and Rymarczyk, Dawid and Struski, {\L}ukasz and Tabor, Jacek and Zieli{\'n}ski, Bartosz},
  booktitle = {Proceedings of the IEEE/CVF Winter Conference on Applications of Computer Vision},
  title     = {Protoseg: Interpretable Semantic Segmentation with Prototypical Parts},
  year      = {2023},
  pages     = {1481--1492},
}

@Misc{Ribeiro2016WhyShouldI,
  author        = {Ribeiro, Marco Tulio and Singh, Sameer and Guestrin, Carlos},
  title         = {"why Should I Trust You?": Explaining the Predictions of Any Classifier},
  year          = {2016},
  archiveprefix = {arXiv},
  doi           = {10.48550/arxiv.1602.04938},
  eprint        = {1602.04938},
  eprinttype    = {arxiv},
  primaryclass  = {cs.LG},
}

@Article{Rudin2019Stopexplainingblack,
  author    = {Rudin, Cynthia},
  journal   = {Nature machine intelligence},
  title     = {Stop Explaining Black Box Machine Learning Models for High Stakes Decisions and Use Interpretable Models Instead},
  year      = {2019},
  number    = {5},
  pages     = {206--215},
  volume    = {1},
  publisher = {Nature Publishing Group UK London},
}

@InProceedings{Li2018Deeplearningcase,
  author    = {Li, Oscar and Liu, Hao and Chen, Chaofan and Rudin, Cynthia},
  booktitle = {Proceedings of the Thirty-Second AAAI Conference on Artificial Intelligence and Thirtieth Innovative Applications of Artificial Intelligence Conference and Eighth AAAI Symposium on Educational Advances in Artificial Intelligence},
  title     = {Deep Learning for Case-Based Reasoning through Prototypes: A Neural Network That Explains Its Predictions},
  year      = {2018},
  address   = {New Orleans, Louisiana, USA},
  publisher = {AAAI Press},
  series    = {AAAI'18/IAAI'18/EAAI'18},
  articleno = {432},
  isbn      = {978-1-57735-800-8},
  numpages  = {8},
}

@InProceedings{Rymarczyk2022Interpretableimageclassification,
  author       = {Rymarczyk, Dawid and Struski, {\L}ukasz and G{\'o}rszczak, Micha{\l} and Lewandowska, Koryna and Tabor, Jacek and Zieli{\'n}ski, Bartosz},
  booktitle    = {European Conference on Computer Vision},
  title        = {Interpretable Image Classification with Differentiable Prototypes Assignment},
  year         = {2022},
  organization = {Springer},
  pages        = {351--368},
}

@Article{Hillen2017MomentsVonMises,
  author   = {Hillen, Thomas and Painter, Kevin J. and Swan, Amanda C. and Murtha, Albert D.},
  journal  = {Mathematical Biosciences and Engineering},
  title    = {Moments of Von Mises and Fisher Distributions and Applications},
  year     = {2017},
  issn     = {1551-0018},
  number   = {3},
  pages    = {673--694},
  volume   = {14},
  doi      = {10.3934/mbe.2017038},
}

@InProceedings{GradCam,
  author    = {Selvaraju, Ramprasaath R. and Cogswell, Michael and Das, Abhishek and Vedantam, Ramakrishna and Parikh, Devi and Batra, Dhruv},
  booktitle = {Proceedings of the IEEE international conference on computer vision},
  title     = {Grad-Cam: Visual Explanations from Deep Networks Via Gradient-Based Localization},
  year      = {2017},
  pages     = {618--626},
}

@Article{Wah2011CaltechUcsdBirds,
  author    = {Wah, Catherine and Branson, Steve and Welinder, Peter and Perona, Pietro and Belongie, Serge},
  title     = {The Caltech-Ucsd Birds-200-2011 Dataset},
  year      = {2011},
  publisher = {California Institute of Technology},
}

@InProceedings{Krause,
  author    = {Krause, Jonathan and Stark, Michael and Deng, Jia and Fei-Fei, Li},
  booktitle = {Proceedings of the IEEE international conference on computer vision workshops},
  title     = {3d Object Representations for Fine-Grained Categorization},
  year      = {2013},
  pages     = {554--561},
}

@Article{Dosovitskiy2020ImageIsWorth,
  author  = {Dosovitskiy, Alexey},
  journal = {arXiv preprint arXiv:2010.11929},
  title   = {An Image Is Worth 16x16 Words: Transformers for Image Recognition at Scale},
  year    = {2020},
}

@Article{Mettes2019HypersphericalPrototypeNetworks,
  author  = {Mettes, Pascal and Van der Pol, Elise and Snoek, Cees},
  journal = {Advances in neural information processing systems},
  title   = {Hyperspherical Prototype Networks},
  year    = {2019},
  volume  = {32},
}

@InProceedings{Nauta2021NeuralPrototypeTrees,
  author    = {Nauta, Meike and van Bree, Ron and Seifert, Christin},
  booktitle = {Proceedings of the IEEE/CVF Conference on Computer Vision and Pattern Recognition (CVPR)},
  title     = {Neural Prototype Trees for Interpretable Fine-Grained Image Recognition},
  year      = {2021},
  month     = jun,
  pages     = {14933--14943},
}

@misc{Wang2024MixtureGaussianDistributed,
  author  = {Wang, Chong and Chen, Yuanhong and Liu, Fengbei and McCarthy, Davis James and Frazer, Helen and Carneiro, Gustavo},
  year={2024},
  eprint={2312.00092},
  archivePrefix={arXiv},
  primaryClass={cs.CV},
  url={https://arxiv.org/abs/2312.00092}, 
}

@Misc{Pach2024LucidPPNUnambiguousPrototypical,
  author        = {Pach, Mateusz and Rymarczyk, Dawid and Lewandowska, Koryna and Tabor, Jacek and Zieliński, Bartosz},
  title         = {Lucidppn: Unambiguous Prototypical Parts Network for User-Centric Interpretable Computer Vision},
  year          = {2024},
  archiveprefix = {arXiv},
  doi           = {10.48550/arxiv.2405.14331},
  eprint        = {2405.14331},
  eprinttype    = {arxiv},
  primaryclass  = {cs.CV},
}

@InProceedings{Sacha2024Interpretabilitybenchmarkevaluating,
  author    = {Sacha, Miko{\l}aj and Jura, Bartosz and Rymarczyk, Dawid and Struski, {\L}ukasz and Tabor, Jacek and Zieli{\'n}ski, Bartosz},
  booktitle = {Proceedings of the AAAI Conference on Artificial Intelligence},
  title     = {Interpretability Benchmark for Evaluating Spatial Misalignment of Prototypical Parts Explanations},
  year      = {2024},
  pages     = {21563--21573},
  volume    = {38},
}

@InProceedings{He2016Deepresiduallearning,
  author    = {He, Kaiming and Zhang, Xiangyu and Ren, Shaoqing and Sun, Jian},
  booktitle = {Proceedings of the IEEE conference on computer vision and pattern recognition},
  title     = {Deep Residual Learning for Image Recognition},
  year      = {2016},
  pages     = {770--778},
}

@InProceedings{Nilsback2008AutomatedFlower,
  author       = "Maria-Elena Nilsback and Andrew Zisserman",
  title        = "Automated Flower Classification over a Large Number of Classes",
  booktitle    = "Indian Conference on Computer Vision, Graphics and Image Processing",
  month        = "Dec",
  year         = "2008",
}

@InProceedings{Wang2021TesNet,
    author    = {Wang, Jiaqi and Liu, Huafeng and Wang, Xinyue and Jing, Liping},
    title     = {Interpretable Image Recognition by Constructing Transparent Embedding Space},
    booktitle = {Proceedings of the IEEE/CVF International Conference on Computer Vision (ICCV)},
    month     = {October},
    year      = {2021},
    pages     = {895-904}
}

@InProceedings{Huang2017Dense,
author = {Huang, Gao and Liu, Zhuang and van der Maaten, Laurens and Weinberger, Kilian Q.},
title = {Densely Connected Convolutional Networks},
booktitle = {Proceedings of the IEEE Conference on Computer Vision and Pattern Recognition (CVPR)},
month = {July},
year = {2017}
}

@inproceedings{Xue2024ProtoPFormer,
  title     = {ProtoPFormer: Concentrating on Prototypical Parts in Vision Transformers for Interpretable Image Recognition},
  author    = {Xue, Mengqi and Huang, Qihan and Zhang, Haofei and Hu, Jingwen and Song, Jie and Song, Mingli and Jin, Canghong},
  booktitle = {Proceedings of the Thirty-Third International Joint Conference on
               Artificial Intelligence, {IJCAI-24}},
  publisher = {International Joint Conferences on Artificial Intelligence Organization},
  editor    = {Kate Larson},
  pages     = {1516--1524},
  year      = {2024},
  month     = {8},
  note      = {Main Track},
  doi       = {10.24963/ijcai.2024/168},
  url       = {https://doi.org/10.24963/ijcai.2024/168},
}

@inproceedings{Vaswani2017Attention,
 author = {Vaswani, Ashish and Shazeer, Noam and Parmar, Niki and Uszkoreit, Jakob and Jones, Llion and Gomez, Aidan N and Kaiser, \L ukasz and Polosukhin, Illia},
 booktitle = {Advances in Neural Information Processing Systems},
 editor = {I. Guyon and U. Von Luxburg and S. Bengio and H. Wallach and R. Fergus and S. Vishwanathan and R. Garnett},
 pages = {},
 publisher = {Curran Associates, Inc.},
 title = {Attention is All you Need},
 url = {https://proceedings.neurips.cc/paper_files/paper/2017/file/3f5ee243547dee91fbd053c1c4a845aa-Paper.pdf},
 volume = {30},
 year = {2017}
}

@misc{chen2017rethinkingatrousconvolutionsemantic,
      title={Rethinking Atrous Convolution for Semantic Image Segmentation}, 
      author={Liang-Chieh Chen and George Papandreou and Florian Schroff and Hartwig Adam},
      year={2017},
      eprint={1706.05587},
      archivePrefix={arXiv},
      primaryClass={cs.CV},
      url={https://arxiv.org/abs/1706.05587}, 
}

@InProceedings{radford21clip,
  title = 	 {Learning Transferable Visual Models From Natural Language Supervision},
  author =       {Radford, Alec and Kim, Jong Wook and Hallacy, Chris and Ramesh, Aditya and Goh, Gabriel and Agarwal, Sandhini and Sastry, Girish and Askell, Amanda and Mishkin, Pamela and Clark, Jack and Krueger, Gretchen and Sutskever, Ilya},
  booktitle = 	 {Proceedings of the 38th International Conference on Machine Learning},
  pages = 	 {8748--8763},
  year = 	 {2021},
  editor = 	 {Meila, Marina and Zhang, Tong},
  volume = 	 {139},
  series = 	 {Proceedings of Machine Learning Research},
  month = 	 {18--24 Jul},
  publisher =    {PMLR},
  url = 	 {https://proceedings.mlr.press/v139/radford21a.html}
}

@inproceedings{wang2023learning,
  title={Learning Support and Trivial Prototypes for Interpretable Image Classification},
  author={Wang, Chong and Liu, Yuyuan and Chen, Yuanhong and Liu, Fengbei and Tian, Yu and McCarthy, Davis J and Frazer, Helen and Carneiro, Gustavo},
  booktitle={Proceedings of the IEEE/CVF International Conference on Computer Vision},
  pages={2062--2072},
  year={2023}
}

@inproceedings{
ma2023this,
title={This Looks Like Those: Illuminating Prototypical Concepts Using Multiple Visualizations},
author={Chiyu Ma and Brandon Zhao and Chaofan Chen and Cynthia Rudin},
booktitle={Thirty-seventh Conference on Neural Information Processing Systems},
year={2023},
url={https://openreview.net/forum?id=dCAk9VlegR}
}

@inproceedings{Kingma2015Adam,
  author       = {Diederik P. Kingma and
                  Jimmy Ba},
  editor       = {Yoshua Bengio and
                  Yann LeCun},
  title        = {Adam: {A} Method for Stochastic Optimization},
  booktitle    = {3rd International Conference on Learning Representations, {ICLR} 2015,
                  San Diego, CA, USA, May 7-9, 2015, Conference Track Proceedings},
  year         = {2015},
  url          = {http://arxiv.org/abs/1412.6980},
  bibsource    = {dblp computer science bibliography, https://dblp.org}
}

@misc{wandb,
title = {Experiment Tracking with Weights and Biases},
year = {2020},
note = {Software available from wandb.com},
url={https://www.wandb.com/},
author = {Biewald, Lukas},
}
